\documentclass[10pt,journal,compsoc]{IEEEtran}

\usepackage{times}
\usepackage{graphicx}
\usepackage{amsmath}   
\usepackage{amssymb}
\usepackage{epstopdf}
\usepackage[hyphens]{url}
\usepackage{subfigure}
\usepackage{enumitem}
\usepackage{algorithm} 
\usepackage{algpseudocode}
\DeclareMathOperator*{\argmin}{arg\,min}
\usepackage{subfigure}
\usepackage{multirow}
\usepackage{dsfont}
\usepackage{booktabs}
\usepackage{colortbl}
\usepackage{tabulary}
\renewcommand\IEEEkeywordsname{Keywords}

\begin{document}



\title{Segment-based Methods for Facial Attribute Detection from Partial Faces}

\author{Upal Mahbub$^{*}$\thanks{* First two authors contributed equally} \quad Sayantan Sarkar$^{*}$   \quad Rama Chellappa\\ 
Department of Electrical and Computer Engineering and the Center for Automation Research, \\UMIACS, University of Maryland, College Park, MD 20742\\
\footnotetext{footnote with two references}
{\tt\small \{umahbub, ssarkar2, rama\}@umiacs.umd.edu}}

\IEEEoverridecommandlockouts

\maketitle

\begin{abstract}
State-of-the-art methods of attribute detection from faces almost always assume the presence of a full, unoccluded face. Hence, their performance degrades for partially visible and occluded faces. In this paper, we introduce SPLITFACE, a deep convolutional neural network-based method that is explicitly designed to perform attribute detection in partially occluded faces. Taking several facial segments and the full face as input, the proposed method takes a data driven approach to determine which attributes are localized in which facial segments. The unique architecture of the network allows each attribute to be predicted by multiple segments, which permits the implementation of committee machine techniques for combining local and global decisions to boost performance. With access to segment-based predictions, SPLITFACE can predict well those attributes which are localized in the visible parts of the face, without having to rely on the presence of the whole face. We use the CelebA and LFWA facial attribute datasets for standard evaluations. We also modify both datasets, to occlude the faces, so that we can evaluate the performance of attribute detection algorithms on partial faces. Our evaluation shows that SPLITFACE significantly outperforms other recent methods especially for partial faces.
\end{abstract}


\section{Introduction and Motivation}
The problem of attribute detection from face images has received much attention from the computer vision community in recent years \cite{JointFaceAndAttribDetect_He_2017} \cite{MOON_Rudd2016} \cite{EmilyNet_AAI_Multitask} \cite{HuHan_DeepMultiTaskHeteroAttrib}. Successful detection of facial attributes has numerous practical applications, such as user-verification \cite{KumarBelhumer_FaceVerificationFromAttribute2009} and image search \cite{KumarBelhumeurImagesearch2011}, video surveillance \cite{videoSurveillance2009}, age and gender estimation to assist salutation for HCI \cite{EmilyNet_AAI_Multitask}, and facial expression estimation for mood analysis \cite{EmilyNet_AAI_Multitask}.
Most attribute detection algorithms assume the availability of a full, near frontal and aligned face, and we find that their performance degrades significantly in domains where partially visible faces are frequent. One such domain is front-camera images of smartphones, which are used for continuous active authentication of users \cite{VMP_SPM_AA_2016} \cite{SegFaceDeepSegFace_FG2017}.

To develop a method that detects attributes from full as well as partial faces, we consider the following key observations: 
\begin{itemize}
\item Some attributes can be inferred correctly even if the face is partially occluded. For example, it is possible for humans to infer the gender from only the left half or upper half of the face. 
\item Some attributes are strongly localized in certain part of the face, such as beard or mustache can only be inferred from the lower half of the face. 
\end{itemize}
Given this observations, it is desirable that a technique for attribute detection be designed, whose performance degrades gracefully with increasing occlusion, rather than suffer catastrophic failures.

In this paper, we present a two-step deep convolutional neural network-based method for facial attribute detection that takes into account the relative strength of different facial segments in detecting different facial attributes. We analyze the detection results obtained in the first step where all facial segments were tasked to decide the attributes. We then present a method to automatically assign selective sets of attributes to different facial regions, resulting in a performance boost in the second step. We also determine the appropriate thresholds for deciding on each attribute at each segment based on the detection results obtained from the validation set. Finally, we combine the predictions from different facial segments and produce the final result. Some special features of the proposed algorithm are:

\begin{itemize}
\item We have implemented a local to global attribute detection approach that harnesses the strength of different facial segments into determining different attributes. For example, bottom-half of a face has information about the beard, while the upper-half has information about the hair. Our divide and conquer approach extracts intermediate results from each segment and combines them in the end to boost the overall performance.

\item Not all facial segments have to be present for the proposed method to work. The method relies on the whole face and one or more facial segments to estimate all the attributes. The individual facial segments are self-sufficient for estimating the attributes they are assigned to. Hence, the method demonstrates superior performance when the full face is not visible due to partial occlusion or pose variation.
 
\item We analyze the local aspects of facial attributes by associating them with facial segments and develop an automated method to utilized the local information.

\item It is a well known fact that an ensemble of networks generally outperform a single network. However training an ensemble is very time-consuming. Our proposed architecture provides us with $16$ predictors with only one round of training. We show that a significant increase in the final result is achieved by combining scores from these predictors.
\end{itemize}

In section \ref{RelatedWorks}, a summary of related works done on facial attribute detection is given. In sections \ref{ProposedMethod}, the proposed Segmentwise, Partial, Localized Inference in Training Facial Attribute Classification Ensembles network is described in detail. All the analyses and experimental results for the proposed methods and comparisons with state-of-the-art methods are provided in section \ref{ExperimentalResults}. Finally, a brief summary of this work as well as future directions of research are included in section \ref{Conclusion}.

\section{Related Works}\label{RelatedWorks}
There has been significant amount of research on attribute extraction starting from learning separate models for each attribute \cite{Huo2016DeepAD} \cite{AgeAndGender_Eidinger} to jointly learning multiple attributes in a multi-task learning fashion \cite{JoingAgeGenderEthnicity_Guo_FGNonCNN} \cite{MOON_Rudd2016} \cite{EmilyNet_AAI_Multitask} \cite{JointFaceAndAttribDetect_He_2017} \cite{HuHan_DeepMultiTaskHeteroAttrib}. Multi-task optimization is found to improve performance in comparison to training independent models for each attribute detection task\cite{EmilyNet_AAI_Multitask} \cite{MOON_Rudd2016}. 

In recent times, the research on attribute detection mostly revolves around two challenging, publicly available datasets namely, CelebA and LFWA \cite{CelebA_liu2015faceattributes}. Both datasets have annotations for forty different attributes along with identity information. The CelebA dataset contains $162,770$ images for training, $19,867$ image for validation and $19,962$ more for testing. It is a very challenging dataset with wide variations in pose, illumination and image quality. The LFWA dataset is a much smaller dataset with $6263$ training and $6880$ test images. The datasets are introduced in \cite{CelebA_liu2015faceattributes} where the authors proposed a cascaded system of two DCNNs to jointly perform face localization and attribute detection. 
In \cite{MOON_Rudd2016} the authors addressed the multi-label imbalance problem of the CelebA dataset and proposed a mixed objective optimization network (MOON) that utilizes a unique loss function comprised of a mixed multitask objective with domain adaptive re-weighting.

Some authors, such as in \cite{EmilyNet_AAI_Multitask} \cite{HuHan_DeepMultiTaskHeteroAttrib}, categorized the attributes into different groups to take advantage of their mutual relationships. The authors in \cite{EmilyNet_AAI_Multitask} suggested an auxiliary network on top of the multi-task DCNN to further exploit the relationships among the attributes. 
On the other hand, the authors in \cite{HuHan_DeepMultiTaskHeteroAttrib} defined a modified AlexNet with both shared and category-specific feature learning to assist attribute extraction. 

Some researchers also implemented the attribute detection task as an auxiliary task of another task. For example, in \cite{JointFaceAndAttribDetect_He_2017}, the authors proposed a DCNN architecture similar to Faster RCNN \cite{FasterRCNN_NIPS2015} with additional losses for joint detection of face and associated facial attributes without requiring explicit face alignment. However, the method does not address partial face detection, which is a challenging problem in itself \cite{DRUID_Umahbub}. 
Other notable attribute detectors for unaligned face are proposed in \cite{UnalignedFA_HuiDing} and \cite{AFFECT_AlignmentFreeAttrib_Boult}. In \cite{UnalignedFA_HuiDing}, the authors proposed a cascade network to concurrently localize face regions to different attributes and perform attribute classification. While this method might be suitable for attribute extraction from partially visible faces if trained properly, the authors presented no such extension or analysis. Also, the original network is huge, consisting of separate DCNN branches for each of the $40$ attributes and therefore not easily scalable. In \cite{AFFECT_AlignmentFreeAttrib_Boult}, the authors introduced a data augmentation technique to assist attribute detection from unaligned faces. They improved the detection performance by augmenting the test data and combining the results. Even though their reported accuracy using an ensemble network of three ResNets is very good on unaligned faces, the architecture does not incorporate any mechanism for partially visible faces and also require combining scores from $162$ transformations of the test image to achieve the best performance.

\section{Proposed Method}\label{ProposedMethod}
\begin{figure}
\centering
\includegraphics[width = 0.4\textwidth]{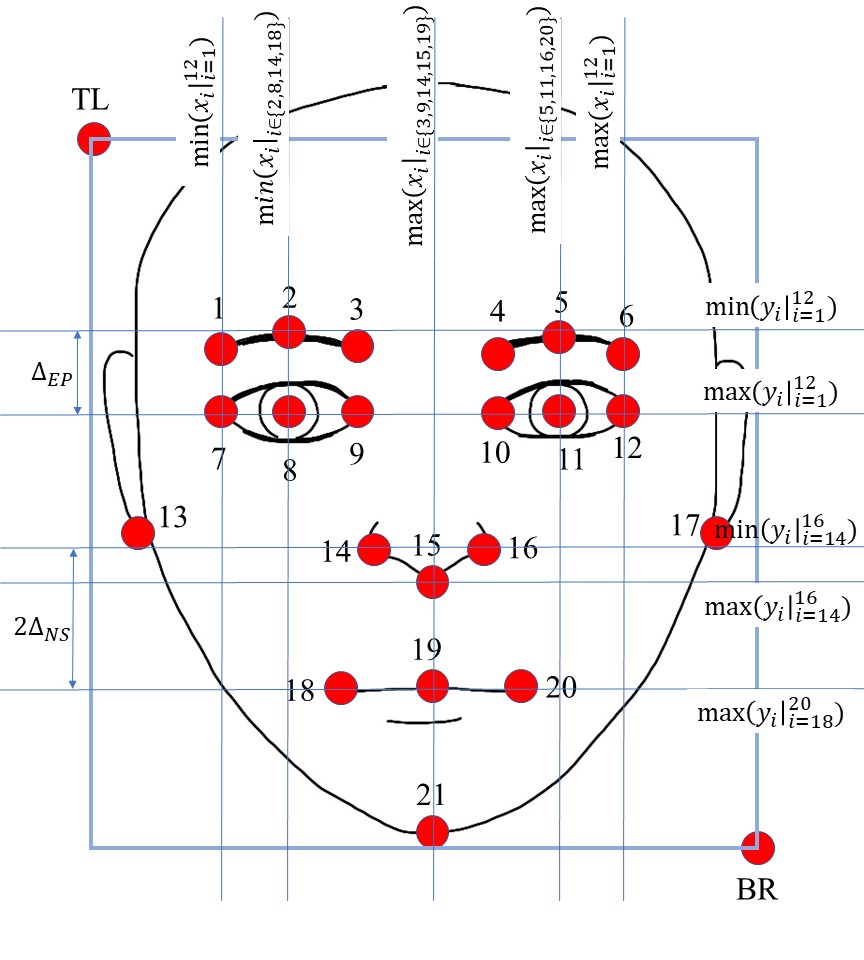}
\caption{The 21 fiducial key points and the full face bounding box.}
\label{Fiducials21}
\end{figure}

The basis of our approach is dividing the task of detecting attributes among different segments of the face. By segments, we mean portions of the face such as left half, right half, upper half, bottom half, nose segment etc. We divide a face into $14$ such facial segments (adopted from \cite{FSFD_Mahbub}) using 21 fiducial keypoints (as shown in Fig. \ref{Fiducials21}) which are upper-left-half (UL12), upper-half (U12), upper-right-half (UR12), upper-left-three-fourth (UL34), upper-three-forth (U34), upper-right-three-fourth (UR34), left-half (L12), left-three-fourth (L34), eye-pair (EP), nose region (NS), right-half (R12), right-three-fourth (R34), bottom-three-fourth (B34), and bottom-half (B12). Let us denote the points shown in Fig. \ref{Fiducials21} with $(x_k, y_k)$ where $x_k$ and $y_k$ are the horizontal and vertical pixel distances from the $(0,0)$ pixel coordinate (top-left corner) of an image image of width $W$ and height $H$, and $k\in\{\mbox{TL, BR},1, 2, \hdots, 21\}$.  TL and BR corresponds to Top-Left and Bottom-Right coordinate of the full face bounding box. The fiducials and full face bounding boxes are obtained from All-in-One Face \cite{UltraFace_RRanjan} along with visibility scores $v_k$ where $k\in\{\mbox{TL, BR},1, 2, \hdots, 21\}$. Now, $\forall v_j\vert_{j=1}^{21}\geq\tau$ where $\tau$ is a visibility threshold, the bounding boxes of segments $\boxdot_L$, were $L\in\{\mbox{UL12, UR12}, \hdots, \mbox{B12}\}$are defined as
\begin{eqnarray}
\boxdot_{\mbox{EP}}&=&[max(x_{\mbox{TL}}, min(x_i\vert _{i=1}^{12})), \nonumber\\ && max(y_{\mbox{TL}}, min(y_i\vert _{i=1}^{12})-\Delta_{\mbox{EP}}),\nonumber\\&& min(x_{\mbox{BR}}, max(x_i\vert _{i=1}^{12})), \nonumber\\&& min(y_{\mbox{BR}}, max(y_i\vert _{i=1}^{12})+\Delta_{\mbox{EP}})]
\nonumber
\end{eqnarray}
\begin{eqnarray}
\boxdot_{\mbox{NS}}&=&[max(x_{\mbox{TL}}, min(x_i\vert_{i\in\{8, 14, 15, 16, 18\}})),\nonumber\\ &&max(y_{\mbox{TL}}, max(0, \frac{1}{3}\sum_{i=14}^{16}y_i - 2\Delta_{\mbox{NS}})),\nonumber\\ &&min(x_{\mbox{BR}}, max(x_i\vert_{i\in\{11, 14, 15, 16, 20\}})) \nonumber\\&& min(y_{\mbox{BR}}, max(H, \frac{1}{3}\sum_{i=14}^{16}y_i + 2\Delta_{\mbox{NS}}))]
\nonumber
\end{eqnarray}
\begin{eqnarray}
\boxdot_{\mbox{UL12}}&=&[x_{\mbox{TL}}, y_{\mbox{TL}}, max(x_i\vert_{i\in\{3, 9, 14, 15, 19\}}), \nonumber\\&&max(y_i\vert_{i=14}^{16})]
\nonumber
\end{eqnarray}
\begin{eqnarray}
\boxdot_{\mbox{U12}}&=&[x_{\mbox{TL}}, y_{\mbox{TL}}, x_{\mbox{BR}}, max(y_i\vert_{i=14}^{16})]
\nonumber
\end{eqnarray}
\begin{eqnarray}
\boxdot_{\mbox{UR12}}&=&[min(x_i\vert_{i\in\{4, 10, 15, 16, 19\}}), y_{\mbox{TL}}, \nonumber\\&& x_{\mbox{BR}}, max(y_i\vert_{i=14}^{16})]
\nonumber
\end{eqnarray}
\begin{eqnarray}
\boxdot_{\mbox{UL34}}&=&[x_{\mbox{TL}}, y_{\mbox{TL}}, max(x_i\vert_{i\in\{5, 11, 16, 20\}}), \nonumber\\&&max(y_i\vert_{i=18}^{20})]
\nonumber
\end{eqnarray}
\begin{eqnarray}
\boxdot_{\mbox{U34}}&=&[x_{\mbox{TL}}, y_{\mbox{TL}}, x_{\mbox{BR}}, max(y_i\vert_{i=18}^{20})]
\nonumber
\end{eqnarray}
\begin{eqnarray}
\boxdot_{\mbox{UR34}}&=&[min(x_i\vert_{i\in\{2, 8, 14, 18\}}), y_{\mbox{TL}}, \nonumber\\&& x_{\mbox{BR}}, max(y_i\vert_{i=18}^{20})]
\nonumber
\end{eqnarray}
\begin{eqnarray}
\boxdot_{\mbox{L12}}&=&[x_{\mbox{TL}}, y_{\mbox{TL}}, max(x_i\vert_{i\in\{3, 15, 19\}}), y_{\mbox{BR}}]
\nonumber
\end{eqnarray}
\begin{eqnarray}
\boxdot_{\mbox{L34}}&=&[x_{\mbox{TL}}, y_{\mbox{TL}}, max(x_i\vert_{i\in\{5, 11, 16, 20\}}), y_{\mbox{BR}}]
\nonumber
\end{eqnarray}
\begin{eqnarray}
\boxdot_{\mbox{R34}}&=&[min(x_i\vert_{i\in\{2, 8, 14, 18\}}), y_{\mbox{TL}}, x_{\mbox{BR}}, y_{\mbox{BR}}]
\nonumber
\end{eqnarray}
\begin{eqnarray}
\boxdot_{\mbox{R12}}&=&[min(x_i\vert_{i\in\{4, 10, 15, 16, 19\}}), y_{\mbox{TL}}, x_{\mbox{BR}}, y_{\mbox{BR}}]
\nonumber
\end{eqnarray}
\begin{eqnarray}
\boxdot_{\mbox{B12}}&=&[x_{\mbox{TL}}, min(y_i\vert_{i=14}^{16}), x_{\mbox{BR}}, y_{\mbox{BR}}]
\nonumber
\end{eqnarray}
\begin{eqnarray}
\boxdot_{\mbox{B34}}&=&[x_{\mbox{TL}}, min(y_i\vert_{i=7}^{12}), x_{\mbox{BR}}, y_{\mbox{BR}}].
\end{eqnarray}
Here, $\Delta_{\mbox{EP}}=max(\vert y_i - y_{i-6}\vert_{i=7}^{12})$ and 
$\Delta_{\mbox{NS}}=0.5*(max(y_i|_{i=18}^{20}) -min(y_i|_{i=14}^{16}))$, $\forall v_j\geq\tau$, where $j\in\{1, 2, \hdots 21\}$.

Intuitively, certain segments are more effective at predicting a subset of attributes than others. For example, we can expect that segments related to the upper part of the face (e.g. $U12$, $U34$ etc.) would contain information about the person being bald or having certain types and color of hair. Therefore, even if the some other part of the face is occluded (e.g. $B12$, $EP$, or $NS$ being not visible), by looking at the upper portion of the face, one can still predict attributes related to hair. Thus detecting attributes from parts as opposed to the whole face, has the advantage of allowing graceful degradation of performance rather than catastrophic failures with increasing occlusion. 

While some attributes can be easily predicted from facial segments, some attributes reflect more global characteristics. For example, one can get hints if a person is young or not from multiple parts of the face, but youth is a global attribute. Therefore it is important to combine the segment predictions into a global prediction, so that multiple parts can contribute to the final prediction. Naturally, the following questions arise:
\begin{itemize}
\item  \emph{Global vs local attributes:}  How does one decide if an attribute is better predicted by facial segments or by a global predictor?
\item \emph{Optimal segment selection:} How does one decide which facial segment is more suitable for predicting a particular attribute?

\item \emph{Combining results multiple networks:} Given that each attribute is predicted by multiple segments of the proposed network, how does one combine the results optimally?

\item \emph{Handling occlusion:} If a certain facial segment responsible for predicting a certain attribute is not visible, how does one get a reasonable prediction?
\end{itemize}


We can summarize the solutions to these problems as follows. 
\begin{itemize}
\item \emph{Network architecture:} The first problem is solved by choosing an architecture of a DCNN that is not only able to predict attributes from facial segments, but also performs feature-level fusion of intermediate features through a  global prediction network to produce accurate global predictions. Also, the sub-modules of the network have a Global Average Pooling which endows the networks with localization ability \cite{Zhou_2016_CVPR}.
\item \emph{Output pruning:} The second question is answered by the two-stage training approach that is adopted in this work, where the first stage primarily is used to prune the outputs of the segment networks by deciding which segments are good at predicting which attributes.
\item \emph{Committee Machines:} For the third problem, we use two-committee machines to perform score level fusion of the multiple predictors to significantly improve the performance of any single constituent network.
\item \emph{Hierarchy of best predictors and Segment Dropout:} Finally, to address the fourth problem, we keep track of a hierarchy of segments, which are good at predicting that attribute. Therefore, even if the best segment for an attribute is not present, one can fall back on other segments that are known to do somewhat well for that attribute. We also train our network with `Segment Dropout' \cite{7961801} to make it more robust to partial faces.

\end{itemize}

These ideas are core to our proposed method: \textbf{S}egmentwise, \textbf{P}artial, \textbf{L}ocalized \textbf{I}nference in \textbf{T}raining \textbf{F}acial \textbf{A}ttribute \textbf{C}lassification \textbf{E}nsembles (SPLITFACE). The algorithm looks at the facial segments and learns to infer local attributes, to better handle partial faces. The next four subsections expand on these ideas.

\subsection{Local to Global Network Architecture}
\begin{figure*}
\centering
\includegraphics[width = 0.95\textwidth]{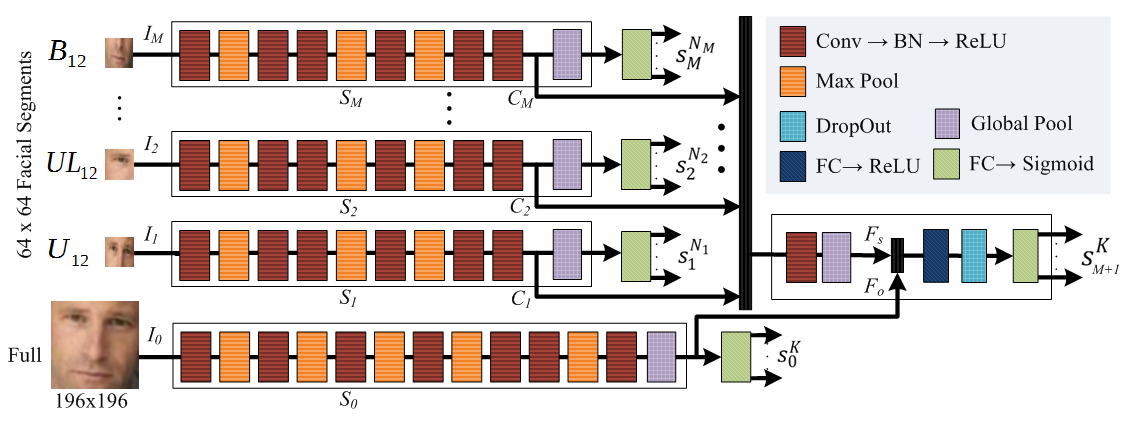}
\caption{SPLITFACE network architecture showing the Facial Segment Networks and the Full Face Network, which are culminate in the Global Prediction Network.}
\label{Task1Diagram}
\end{figure*}

Here the three constituents of the proposed network namely, the Full Face Network, the Facial Segment Networks and the Global Predictor Network, and their training losses are described:

\emph{Facial Segment Networks}: Let $I_1, I_2, \hdots, I_M$ denote face regions for the aforementioned  $M=14$ facial segments. Each segment has some predictive power which is unknown initially. In the next section, we describe our data-driven approach to find which attributes are predicted more accurately by each segment. For now, let us say that segment $i$ predicts a set of attributes $N_i$, where the number of attributes predicted by each segment $|N_i| \le K$, where $i\in\{1, 2, \hdots, M\}$. Initially, all segments predict all $K=40$ attributes, but later each segment is allowed to specialize, as described in the next section. We denote these Segment Networks $S_i$, where $i\in\{1, 2, \hdots, M\}$. When the facial segment $I_i$ is passed through its corresponding segment network $S_i$, it yields attribute scores $s_i$ for each attribute in $N_i$ and a feature for that segment $C_i$, i.e. 
\begin{equation}
s_i, C_i = S_i(I_i)
\label{segeq}
\end{equation}
where $C_i$ is tapped from the last convolutional layer of $S_i$. The architectures of all the segment networks $S_i$ are same, as described in table \ref{NetArchTable}, and each of these segment networks is independent of one another. 

\emph{Full Face Network}: Let $I_0$ represent the full-face region, which is passed through a DCNN $S_0$. We have adopted a seven layer deep convolutional network as $S_0$. Details on the network architecture are provided in table \ref{NetArchTable}. The full face region is expected to always predict all $K$ attributes ($|N_0|=K$). Hence, it outputs a vector $s_0$ of length $K$, and also a compact feature representation $F_0$ after global pooling the last convolutional feature, i.e.
\begin{equation}
s_0, F_0 = S_0(I_0)
\label{fulleq}
\end{equation}

\emph{Global Prediction Network for local feature fusion}: In the Global Prediction Network, we combine the results from the local segment networks and the full face network to produce predictions for all the $K$ attributes. To do so, we first concatenate the convolutional features $C_i$ from the $M$ segment networks, convolve them, then apply global pooling to get a flattened feature from the segments, $F_s$. This is concatenated with $F_0$, the flattened features from the Full Face Network, and passed through a few fully connected layers, to finally yield predictions for all the $K=40$ attributes. The Global Prediction Network can be thought of as a feature level fusion of the different segments, as opposed to the score level fusion of committee machines described in the next section.

\begin{equation}
s_{M+1} = S_{M+1}(F_0, C_1, \hdots, C_M)
\label{globeq}
\end{equation}

The color-coded network architecture for the SPLITFACE network is shown in Fig. \ref{Task1Diagram}. It shows the above mentioned architectural choices, namely predictions from segments, predictions from the full face and fusion of segment and full face features to provide a global prediction.

For further discussions, we shall use the word `predictor' to mean any of the $M+2$ sub-networks $S_i, i \in \{0,1,\hdots,M+1\}$ described above, that is, any of the $M$ Facial Segment Networks, the Full Face Network or the Global Prediction Network.

\emph{Localization using Global Average Pooling}: It has been shown in \cite{Zhou_2016_CVPR} that Global Average Pooling (GAP) introduced in \cite{NIN} has remarkable localization properties. Since we are aiming to predict localized attributes well from partial segments, we use a GAP layer in the architecture to transition from convolutional to fully connected layers. Using Class Activation Maps (CAM) in section \ref{camheatmap} we observe that this provides the network with the desired property of being able to focus on regions of interest, thus making the process more interpretable.

\emph{Loss}: We use binary crossentropy loss for all the predictor outputs $s_i, i \in \{0,1,\hdots,M+1\}$ described in \ref{segeq}, \ref{fulleq}, \ref{globeq}, weighted by the inverse of priors. Then loss $L$ incurred on image $I$ is:
\begin{equation}
L(I) = \sum_{i=0}^{M+1} \sum_{j=0}^{K-1} w_{I,j} \log(s_i(j))
\label{losseq}
\end{equation}
In \ref{losseq}, $w_{I,j}$ is a weight based on the ground truth $I_j$ and the prior probability $p_j$ of attribute $j$ being present, which is precomputed on the training set. The weight $w_{I,j}$ defined in \ref{priorweight} helps to mitigate the challenges due to unbalanced class distributions which are prevalent in datasets like CelebA \cite{MOON_Rudd2016}.
\begin{equation}
w_{I,j}=\begin{cases}
  p_j,& \text{if } I_j=0 \\
  1-p_j & \text{if } I_j=1
\end{cases}
\label{priorweight}
\end{equation}



\begin{table*}
\centering
\caption{Detailed network architecture.}
\begin{tabular}{c | c | c}
\hline
$S_i$, $i\in\{1,2, \hdots, M\}$						& $S_0$													& $S_{M+1}$	\\
\hline
\hline
conv$3$-$3$2 $\rightarrow$ BN $\rightarrow$	ReLU	& conv$3$-$32$ $\rightarrow$ BN $\rightarrow$	ReLU 	& conv$3$-$512$ $\rightarrow$ BN $\rightarrow$	ReLU	\\
\hline
2D MaxPool $3\times 3$, Stride $2$					& 2D MaxPool $3\times 3$, Stride $2$		& global average pooling\\
\hline
conv$3$-$64$ $\rightarrow$ BN $\rightarrow$	ReLU 	& conv$3$-$64$ $\rightarrow$ BN $\rightarrow$	ReLU	& merge$(F_0, F_s)$\\
\hline
conv$3$-$64$ $\rightarrow$ BN $\rightarrow$	ReLU 	& 2D MaxPool $3\times 3$, Stride $2$	&dense-$256$D$\rightarrow$ ReLU\\
\hline
2D MaxPool $3\times 3$, Stride $2$ 					& conv$3$-$64$ $\rightarrow$ BN $\rightarrow$	ReLU	&	dropout$(0.2)$\\
\hline
conv$3$-$128$ $\rightarrow$ BN $\rightarrow$ ReLU	& 2D MaxPool $3\times 3$, Stride $2$	& \\
\hline
2D MaxPool $3\times 3$, Stride $2$					& conv$3$-$128$ $\rightarrow$ BN $\rightarrow$ ReLU		& \\
\hline
conv$3$-$128$ $\rightarrow$ BN $\rightarrow$ ReLU	& 2D MaxPool $3\times 3$, Stride $2$\\
\hline
conv$3$-$256$ $\rightarrow$ BN $\rightarrow$ ReLU	& conv$3$-$128$ $\rightarrow$ BN $\rightarrow$ ReLU	& \\
\hline
													& conv$3$-$256$ $\rightarrow$ BN $\rightarrow$ ReLU	& \\
\hline
													& 2D MaxPool $3\times 3$, Stride $2$	& \\
\hline
													& conv$3$-$256$ $\rightarrow$ BN $\rightarrow$ ReLU	& \\
\hline
\end{tabular}
\label{NetArchTable}
\vskip -5pt
\end{table*}


\subsection{Optimal segment selection for output pruning}
Intuitively, not all segments predict all attributes well. Therefore it is counterproductive to train the network to produce all $K$ predictions from all $M$ segment networks. Instead, we follow a data-driven approach to prune the segment networks.

\emph{Stage 1}: Initially, we predict all $K=40$ attributes from all $M$ segment networks, the full face network and the global prediction network. Hence each attribute is predicted by $M+2$ networks. After training for several epochs, we evaluate the detection accuracy of each of the $M$ segment networks, $S_i$. For each attribute, we sort the $M+2$ networks according to accuracy on validation set, and pick the top $d$ networks for each attribute. The global predictor (GP) and the full face network can be expected to be among the best predictors most of the time, since they have a top view of the sum of parts. The rest of the $d-2$ predictors of each attribute are segment networks and therefore the most associated $N_i$ attributes for segment $i$ where $i\in\{1,2, \hdots, M\}$ and $N_i\leq K$ is determined this way.

\begin{figure*}
\centering
\includegraphics[width = 0.98\textwidth]{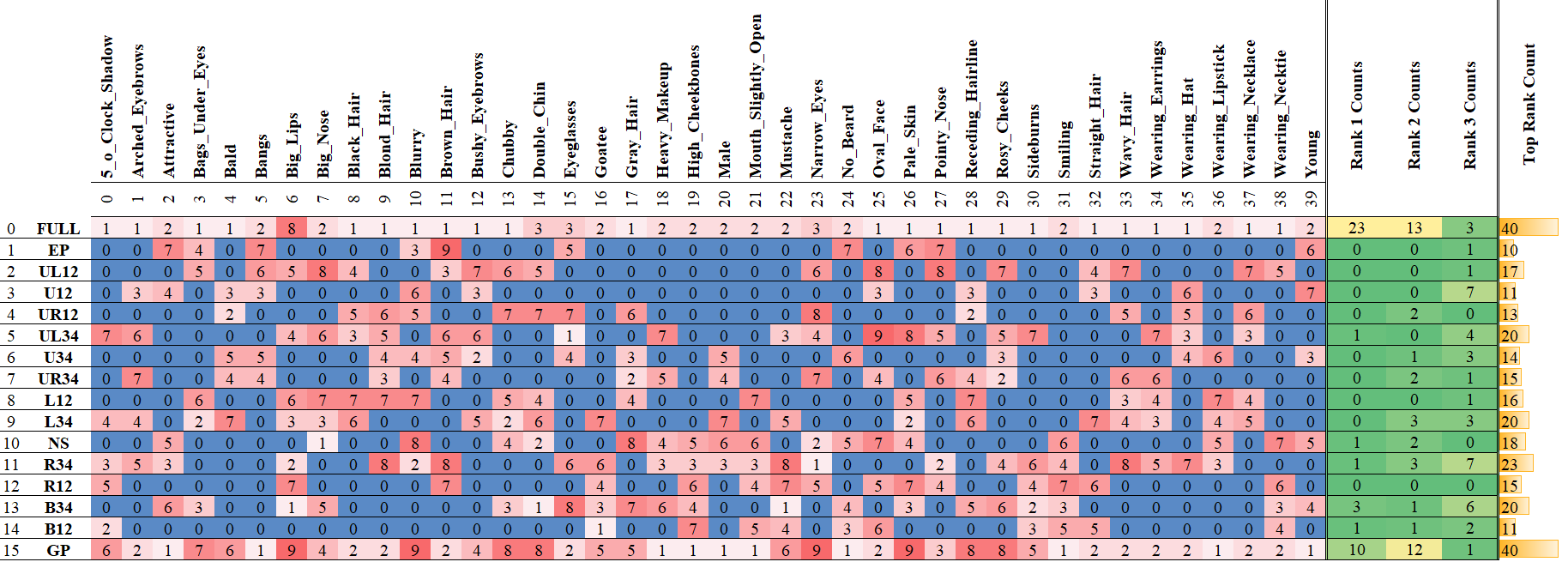}
\caption{The top ranked segments (including GP and Full face, row-wise) for each attribute (in the columns). The blue cells indicate that that particular segment was not used to predict that attribute in the stage $2$ of training. The segments predict attributes that are localized in that region. For example, the bottom half segment predicts attributes related to facial hair.}
\label{Top7SegHeatMap}
\end{figure*}

In Fig.\ref{Top7SegHeatMap} a table for $d=7$ and $M=14$ is shown where for the segment networks (row three and below) the non-zero numbers for different attribute columns denote the attributes assigned to the segment after pruning. The total number of attributes assigned to the segment networks after pruning are shown in the last column.

\emph{Stage 2:} After the association of attributes with segments as described above, a second round of training is performed. The pruning process in stage 1 allows the segment networks to focus on attributes that they perform best on, without having to worry about attributes they are just not capable of predicting. Also, we have intentionally assigned all the attributes to GP and FULL networks (as shown in Fig. \ref{Top7SegHeatMap}), since the receptive field for these two networks encompasses the entire face. So, we make the inherent assumption that these two networks are capable of successfully predicting all the attributes.


\subsection{Committee Machines for Score-Level Fusion}

While the global predictor or the full face networks has good predictive power, using only those two predictors does not harness the full potential of the proposed network architecture. To utilize all the segment networks along with the global and full face predictors, we describe two committee machines here namely, the Highest Ranked Predictor (HRP) and the Normalized Score Aggregation (NSA) methods, that perform score-level fusion. For both methods, using the validation set $V$, we first compute the optimal thresholds, $t_{a,i_a}$ for each attribute $a\in \{1,2,\hdots,K\}$ and for each predictor $i_a\subset \{0, 1, \hdots, M+1\}$ which are responsible for predicting $a$. For CelebA, the table in Fig. \ref{Top7SegHeatMap} shows information about $i_a$, which is an ordered set or tuple, ordered in descending order of validation accuracy. For example, if we consider the attribute `goatee', then $a=16$ and $i_{a=16}=(15, 1, 14, 13, 0, 12, 10)$, which correspond to B12, FULL, B34, R12, GP, R34 and L34. Denoting $I_a$ as the ground truth of attribute $a$ for sample $I$ and $\mathds{1}$ as the indicator function, the optimal thresholds that maximize validation accuracy are computed as:
\begin{equation}
t_{a,i} = \argmin_{t \in [0,1]} \sum^{}_{I\in V} \mathds{1}_{I_a = \mathds{1}_{S_i(I)<t}}
\label{optth}
\end{equation}

We denote visibility of a segment $j$ for image $I$ by $V_{I,j}\in\{0,1\}$. Clearly $V_{I,0}=V_{I,15}=1$, since the both the Full Face Predictor and Global Predictor Network can predict attributes no matter what the occlusion is. Finally, we define an ordered set $i_a^v$ which contains the top usable predictors for attribute $a$ (ones that have visible segments) as
\begin{equation}
i_a^v = (x : x \in i_a, V_{I,x}=1).
\end{equation}


\subsubsection{Highest Ranked Predictor (HRP) Committee Machine}
After the completion of the two training stages, we evaluate the performance of each of the predictors (segment, full face and global networks) on the validation set to find a hierarchy of best performing predictors for each attribute. The results are shown in the table in Fig. \ref{Top7SegHeatMap}. For example, we can see that the best predictors for `goatee' are B12, FULL, B34, R12, GP, R34 and L34, in descending order of validation accuracy. 

When making a prediction for an attribute, we find the topmost usable predictor $j=i_a^v(1)$ that is usable/visible for that image. This score from predictor $S_j(I_j)$ is then thresholded with the optimal threshold of that segment for that attribute, which was precomputed from the validation set following \ref{optth}. The prediction outcome based on the optimal threshold would be
\begin{equation}
P_a(I) = \mathds{1}_{S_j(I_j)<t_{a,j}}.
\label{v2eqn}
\end{equation}
Continuing our example, to predict `goatee', we use the prediction of B12 ($=i_a(0)$), and if that segment is not visible, we use FULL ($=i_a(1)$). 


\subsubsection{Normalized Score Aggregation (NSA) Committee Machine}
In general, different predictors trying to predict the same attribute might have different optimal thresholds. Once they are aggregated (say, by taking their mean or product or median), one needs to calculate the optimal threshold for the aggregate score. Instead, we could normalize the scores of the predictors so that after aggregation, the optimal threshold for the aggregate score is $0.5$. \cite{Cappelli_ThresholdNorm} suggests a double sigmoid score normalization function for fusing scores from multiple predictors. However, it involves $2$ hyper-parameters, which need to be found by cross-validation. Instead we propose a simpler normalization function below, which does not require any hyper-parameters.

\emph{Linear Threshold Normalization: } Consider a binary classification problem, where we have to decide the class $C\in \{0,1\}$, given a score $X\in [0,1]$. We assume that the optimal threshold $t$ that maximizes separation between the 2 classes is known. Therefore, $P(C=1|X=t)=0.5$. Consider a transformation $Y=T(X)$. We wish to identify the function $T$, such that, $P(C=1|Y=0.5)=0.5$.

\begin{eqnarray}
P(C=1|Y=0.5)&=&P(C=1|T(X)=0.5)\nonumber\\
&=& P(C=1|X=T^{-1}(0.5))\nonumber\\
\end{eqnarray}

If we choose an invertible function $T$, such that $T(t)=0.5$, then above equation yields $P(C=1|Y=0.5)=0.5$. Thus, given multiple scores $X_i$, and their optimal thresholds $t_i$, we can transform the scores to $Y_i=T(X_i)$, so that after aggregating $Y_i$, say by averaging, the optimal threshold is $0.5$.

For our algorithm, we use a piecewise linear transformation \ref{piecewiselin}, that satisfies the $T(t)=0.5$ criterion discussed above.
\begin{equation}
T_t(x) = 
\begin{cases}
  (0.5/t) \times x,& \text{if } 0 \leq x \leq t \\
  (0.5x + 0.5 -t)/(1-t) & \text{if } t <x\leq 1
\end{cases}
\label{piecewiselin}
\end{equation}



We transform the scores of at least $p=5$ top predictors out of the visible ones, $i_a^v$, using \ref{piecewiselin} to yield $Z$ (\ref{transformedscores}).
\begin{equation}
Z=\{T_{t_{a,i_a^v(k)}}(S_{i_a^v(k)}(I_{i_a^v(k)})) : k\in\{1,2,\hdots,min(|i_a^v|,p)\}\}
\label{transformedscores}
\end{equation}

Finally we use an aggregation function $A$ on $Z$ for the prediction. The decision rule using the aggregator function is
\begin{equation}
P_a(I) = \mathds{1}_{A(Z) \geq A(\{1-z:z\in Z\})}.
\end{equation}

Possible choices of aggregator functions are:
\begin{itemize}
\item \emph{Bayes' Rule or Product Rule}: As discussed in the score fusion literature \cite{Jain_CombinationApproach} we can use a product rule to combine the decisions of $N$ binary classifiers according to the following decision rule
\begin{equation}
\prod^N P(C=1|x)\lessgtr \prod^N (1-P(C=1|x)).
\label{bayesrule}
\end{equation}
Thus the product aggregator function is $A(Z)=\prod_{z\in Z}z$ for all $z\in Z$.
\item \emph{Median Rule}: As proposed in \cite{DBLP:journals/corr/cs-NE-9905012}, the median aggregator function is $A(Z)=med(Z)$.
\end{itemize}









\subsection{Segment Dropout and Hierarchy of Best Predictors for Handling Occlusion}


\emph{Segment Dropout:} When training the network with image $I$, only a subset of the $M=14$ segments might be present. The visible segments are randomly dropped with probability $30\%$ when training. This is called Segment Dropout, which was introduced in \cite{7961801} to augment the dataset for handling occlusion. When a certain segment is not present in a face the input to corresponding segment branch is zero. In order to make SPLITFACE robust against such cases and generalize better to detect attributes from the available segments, random segment dropout is performed.

\emph{Hierarchy of Best Predictors:} As described in the earlier section, we compute a hierarchy of predictors $i_a^v$ that are visible. Thus even if a face is partially occluded and the best segment is not available, the other segments provide reasonable predictive power to the committee machine.


The unique architecture of SPLITFACE allows the use of predictor hierarchy and thereby improving the detection accuracy. In addition, it ensures that even if some part of the input face is not visible due to occlusion or failure of the face detector, the attribute detection network would rely on the visible segments to still make a good prediction. Note that the input to GP are the features from all the segment networks and the full face network, and our partial face augmentation approach during training enables it to handle missing segments while predicting attributes.

\section{Experimental Setup and Evaluation}\label{ExperimentalResults}
\subsection{Datasets}
\label{datasets}
\begin{figure}
\centering
\includegraphics[width = 0.48\textwidth]{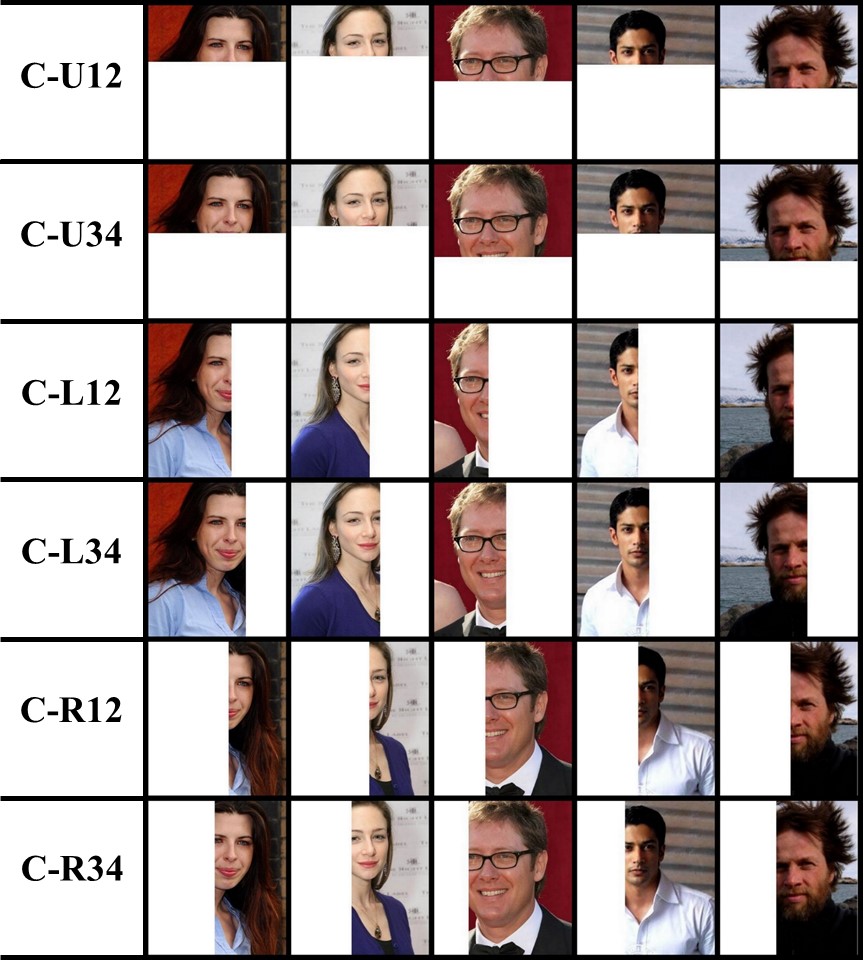}
\caption{Modified CelebA dataset samples for partial faces.}
\label{FSDatasetSamples}
\end{figure}

We use the CelebA \cite{CelebA_liu2015faceattributes} and LFWA \cite{CelebA_liu2015faceattributes} datasets for both training and evaluation. Also, to evaluate the SPLITFACE's capability for handling partially visible faces when estimating facial attributes, we created several variations of these two datasets and evaluate the performance of SPLITFACE on those variations. We follow the data augmentation scheme described in \cite{DRUID_Umahbub} for generating partially visible faces by cropping the images keeping only L12 or L34 or R12 or R34 or U12 or U34 portion. We replace the rest of the pixels with white pixels. Hence, we create six variations of both datasets and named them C$-{P}$ and L$-{P}$, respectively for CelebA and LFWA, where $P\in\{\mbox{L12, L34, R12, R34, U12, U34}\}$ \footnote[1]{Bounding boxes for the partial CelebA and partial LFWA datasets are available at \url{https://drive.google.com/open?id=16hL7g3d6dfvbdvwarYfT6zNcNNXcRLlr}}. Some sample images for C$-{P}$ dataset are shown in Fig. \ref{FSDatasetSamples}.

\subsection{Implementation Details}
The proposed network has $26,090,334$ trainable parameters, which are tuned using the adaptive moment estimation (ADAM) optimizer \cite{ADAMOptimizer}. The initial learning rate was set to $0.001$. For CelebA, we train the network for $10$ epochs for both stages, while for LFWA, we train it for $180$ epochs in the first stage and $270$ epochs in the second stage. The full face region is resized to $196\times 196 \times 3$ and given as input to the full face branch, which the inputs to the facial segment branches are all resized to $64 \times 64 \times 3$. The experiments were performed on NVIDIA Quadro P6000 GPUs, with training batch size of $200$, and the code was written using the KERAS python library \cite{KERAS_chollet2015keras} with tensorflow \cite{Tensorflow} backend. Apart from segment dropout, horizontal flipping was applied for data augmentation. Among state-of-the-art methods, the authors of AFFACT \cite{AFFECT_AlignmentFreeAttrib_Boult} have provided the source code in their paper, which we used for performance comparison on partial face datasets. However, the accuracy obtained from this implementation is slightly less than the accuracy reported in \cite{AFFECT_AlignmentFreeAttrib_Boult}, perhaps because we have not applied test time data augmentation. For all the other methods, we directly report the results in corresponding publications.

\subsection{Visualizing Network Response using Class Activation Maps}
\label{camheatmap}

\begin{figure*}
\centering
\includegraphics[width = \textwidth]{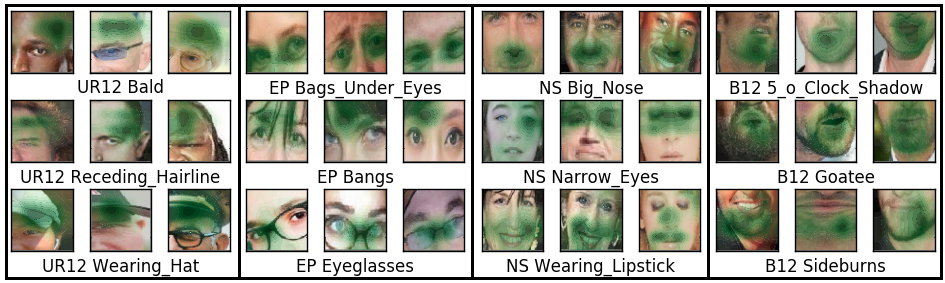}
\caption{Visualization of Class Activation Maps for four different facial segments (UR12, EP, NS and B12 in the four quarters from left to right) and some attributes estimated by the corresponding block.}
\label{CAMheatmap}
\end{figure*}

The class activation map was proposed in \cite{Zhou_2016_CVPR} to visualize the localization properties of the network. Given a network which terminates in a Global Average Pooling (GAP) layer followed by a dense layer, we can compute the CAM $C_i$ of a particular class $i$ as a weighted average of the activation maps of the layer just before the GAP layer as
\begin{equation}
C_i = \sum^N_j w_{i,j}M_j,
\label{CAMeq}
\end{equation}
where $M_j$ is the $j^{\text{th}}$ feature map in a feature tensor of depth $N$ just before the GAP layer and $w_{i,j}$ is the corresponding weight of the dense layer after the GAP layer. In Fig. \ref{CAMheatmap}, we show the CAM superimposed on some facial segments from CelebA. Clearly, the activation maps are localized in interpretably meaningful regions. It can be seen from Fig. \ref{CAMheatmap} that for all the three attributes shown for UR12 (`bald', `receding hairline' and `wearing hat') and B12 (`5 o clock shadow', `goatee' and `sideburns'), the network focuses on the same region: the top corner of the head for UR12 and the chin and cheeks for B12.
On the other hand, for segments EP and NS, the attention shifted to different regions for different attributes. For example for segment EP, the attribute `bags under eyes' is predicted when the network has high response near the eyes, `bangs' are predicted when the response is high near the forehead and `eyeglasses' are predicted by looking at the bridge of the nose. Similarly the NS segment network shifted its attention to the nose, eyes or lips to predict `big nose', `narrow eyes' and `wearing lipstick', respectively.

\subsection{Performance Comparison on Original CelebA and LFWA datasets}
The performance of the proposed method is compared with state-of-the-art methods in tables \ref{tab:PerformanceCompareOnCelebA} and \ref{tab:CompareResultsLFWA} on the original CelebA and LFWA datasets, respectively. Among state-of-the-art methods, the result of AFFACT is directly reported from \cite{AFFECT_AlignmentFreeAttrib_Boult}, while the column AFFACT Unaligned contains results that we found by evaluating the full faces we use in our experiment. Since the source codes were not available for any of the other state-of-the-art methods, we reported the results directly from corresponding publications. The column titled `Prior' shows the accuracies obtainable by only applying the knowledge from the prior probabilities of the presence or absence of an attribute in the datasets. It can be seen that a staggering mean accuracy of $80.57\%$ in CelebA and $71.27\%$ in LFWA is achievable by only using the prior probabilities in decision making. Even though the state-of-the-art methods and the proposed method increases this number by more than $10\%$, for certain attributes such as Big Lips and Narrow Eyes in table \ref{tab:PerformanceCompareOnCelebA}, the prior is higher than the trained methods for most of the approaches. We put the prior column in the table as a baseline for evaluation. 

The last five columns in tables \ref{tab:PerformanceCompareOnCelebA} and \ref{tab:CompareResultsLFWA} show the attribute-wise accuracy and the mean accuracy for Full, GP, HRP, NSA Product rule and NSA Median rule, respectively, in both tables. It can be seen from these tables that the committee machine approaches boost the results obtained from Full and GP for most of the attributes. The mean accuracy of $90.42\%$ for CelebA and $85.85\%$ for LFWA obtained from NSA Product Rule closely matches the state-of-the-art results presented in the table. Note that we adopted a very simple six layer convolutional network for the Full face branch that achieves $90.72\%$ accuracy on CelebA and $84.02\%$ accuracy on LFWA. The result for CelebA is boosted for HRP but degrades slightly for NSA methods. On the other hand, for LFWA, the committee machine approaches improves the overall performance. Since LFWA is a much smaller dataset and hence the trained network over-fitted greatly on the training set, this boost in result shows that the proposed committee machine approaches, especially NSA, is effective for generalization due to their ensemble aggregation mechanism. In later sections, we will present results for partially visible faces, where the committee machine approaches consistently improves over Full and GP branches and hence the adaptation of such methods is justified for practical purposes even with a slight loss in accuracy for the original dataset.

\begin{table*}[htbp]
  \centering
  \caption{Attribute detection performance comparison on the CelebA dataset in terms of individual and mean detection accuracy for the attributes.}
  \scalebox{0.7}{
    \begin{tabular}{p{9.91em}ccccccccccccc}
    \toprule
    \multirow{3}[6]{*}{\textbf{Attributes}} & \cellcolor[rgb]{ .988,  .894,  .839} & \cellcolor[rgb]{ 1,  .949,  .8} & \cellcolor[rgb]{ 1,  .949,  .8} & \cellcolor[rgb]{ 1,  .949,  .8} & \cellcolor[rgb]{ 1,  .949,  .8} & \cellcolor[rgb]{ 1,  .949,  .8} & \cellcolor[rgb]{ 1,  .949,  .8} & \cellcolor[rgb]{ 1,  .949,  .8} & \cellcolor[rgb]{ .886,  .937,  .855}\textcolor[rgb]{ .886,  .937,  .855}{}Proposed & \cellcolor[rgb]{ .886,  .937,  .855}\textcolor[rgb]{ .886,  .937,  .855}{}Proposed & \multicolumn{3}{c}{\cellcolor[rgb]{ .776,  .878,  .706}\textbf{Proposed Committee Machine }} \\
\cmidrule{12-14}    \multicolumn{1}{c}{} & \multicolumn{1}{c}{\cellcolor[rgb]{ .988,  .894,  .839}\textbf{Prior}} & \multicolumn{1}{c}{\cellcolor[rgb]{ 1,  .949,  .8}\textbf{LENet+}} & \multicolumn{1}{c}{\cellcolor[rgb]{ 1,  .949,  .8}\textbf{MOON\cite{MOON_Rudd2016}}} & \multicolumn{1}{c}{\cellcolor[rgb]{ 1,  .949,  .8}\textbf{MCNN+}} & \multicolumn{1}{c}{\cellcolor[rgb]{ 1,  .949,  .8}\textbf{DMTL\cite{HuHan_DeepMultiTaskHeteroAttrib}}} & \multicolumn{1}{c}{\cellcolor[rgb]{ 1,  .949,  .8}\textbf{AFFACT\cite{AFFECT_AlignmentFreeAttrib_Boult}}} & \multicolumn{1}{c}{\cellcolor[rgb]{ 1,  .949,  .8}\textbf{AFFACT}} & \multicolumn{1}{c}{\cellcolor[rgb]{ 1,  .949,  .8}\textbf{PaW \cite{UnalignedFA_HuiDing}}} & \multicolumn{1}{c}{\cellcolor[rgb]{ .886,  .937,  .855}\textbf{FULL}} & \multicolumn{1}{c}{\cellcolor[rgb]{ .886,  .937,  .855}\textbf{GP}} & \multicolumn{1}{c}{\cellcolor[rgb]{ .776,  .878,  .706}\textbf{ HRP  }} & \multicolumn{2}{c}{\cellcolor[rgb]{ .776,  .878,  .706}\textbf{NSA}} \\
\cmidrule{13-14}    \multicolumn{1}{c}{} & \cellcolor[rgb]{ .988,  .894,  .839} & \multicolumn{1}{c}{\cellcolor[rgb]{ 1,  .949,  .8}\textbf{Anet\cite{CelebA_liu2015faceattributes}}} & \cellcolor[rgb]{ 1,  .949,  .8} & \multicolumn{1}{c}{\cellcolor[rgb]{ 1,  .949,  .8}\textbf{AUX\cite{EmilyNet_AAI_Multitask}}} & \cellcolor[rgb]{ 1,  .949,  .8} & \cellcolor[rgb]{ 1,  .949,  .8} & \multicolumn{1}{c}{\cellcolor[rgb]{ 1,  .949,  .8}\textbf{Unaligned\cite{AFFECT_AlignmentFreeAttrib_Boult}}} & \cellcolor[rgb]{ 1,  .949,  .8} & \cellcolor[rgb]{ .886,  .937,  .855} & \cellcolor[rgb]{ .886,  .937,  .855} & \multicolumn{1}{c}{\cellcolor[rgb]{ .776,  .878,  .706}\textbf{}} & \multicolumn{1}{c}{\cellcolor[rgb]{ .776,  .878,  .706}\textbf{Prod. Rule}} & \multicolumn{1}{c}{\cellcolor[rgb]{ .776,  .878,  .706}\textbf{Med. Rule}} \\
    \midrule
    \textbf{5\_o\_Clock\_Shadow} & \cellcolor[rgb]{ .988,  .894,  .839}88.83 & \cellcolor[rgb]{ 1,  .949,  .8}91.00 & \cellcolor[rgb]{ 1,  .949,  .8}94.03 & \cellcolor[rgb]{ 1,  .949,  .8}94.51 & \cellcolor[rgb]{ 1,  .949,  .8}95.00 & \cellcolor[rgb]{ 1,  .949,  .8}94.21 & \cellcolor[rgb]{ 1,  .949,  .8}94.09 & \cellcolor[rgb]{ 1,  .949,  .8}94.64 & \cellcolor[rgb]{ .886,  .937,  .855}93.96 & \cellcolor[rgb]{ .886,  .937,  .855}90.00 & \cellcolor[rgb]{ .776,  .878,  .706}93.96 & \cellcolor[rgb]{ .776,  .878,  .706}93.01 & \cellcolor[rgb]{ .776,  .878,  .706}93.13 \\
    \midrule
    \textbf{Arched\_Eyebrows} & \cellcolor[rgb]{ .988,  .894,  .839}73.41 & \cellcolor[rgb]{ 1,  .949,  .8}79.00 & \cellcolor[rgb]{ 1,  .949,  .8}82.26 & \cellcolor[rgb]{ 1,  .949,  .8}83.42 & \cellcolor[rgb]{ 1,  .949,  .8}86.00 & \cellcolor[rgb]{ 1,  .949,  .8}82.12 & \cellcolor[rgb]{ 1,  .949,  .8}81.27 & \cellcolor[rgb]{ 1,  .949,  .8}83.01 & \cellcolor[rgb]{ .886,  .937,  .855}83.39 & \cellcolor[rgb]{ .886,  .937,  .855}83.44 & \cellcolor[rgb]{ .776,  .878,  .706}83.39 & \cellcolor[rgb]{ .776,  .878,  .706}82.44 & \cellcolor[rgb]{ .776,  .878,  .706}82.56 \\
    \midrule
    \textbf{Attractive} & \cellcolor[rgb]{ .988,  .894,  .839}51.36 & \cellcolor[rgb]{ 1,  .949,  .8}81.00 & \cellcolor[rgb]{ 1,  .949,  .8}81.67 & \cellcolor[rgb]{ 1,  .949,  .8}83.06 & \cellcolor[rgb]{ 1,  .949,  .8}85.00 & \cellcolor[rgb]{ 1,  .949,  .8}82.83 & \cellcolor[rgb]{ 1,  .949,  .8}80.36 & \cellcolor[rgb]{ 1,  .949,  .8}82.86 & \cellcolor[rgb]{ .886,  .937,  .855}82.71 & \cellcolor[rgb]{ .886,  .937,  .855}82.86 & \cellcolor[rgb]{ .776,  .878,  .706}82.86 & \cellcolor[rgb]{ .776,  .878,  .706}83.13 & \cellcolor[rgb]{ .776,  .878,  .706}82.76 \\
    \midrule
    \textbf{Bags\_Under\_Eyes} & \cellcolor[rgb]{ .988,  .894,  .839}79.55 & \cellcolor[rgb]{ 1,  .949,  .8}79.00 & \cellcolor[rgb]{ 1,  .949,  .8}84.92 & \cellcolor[rgb]{ 1,  .949,  .8}84.92 & \cellcolor[rgb]{ 1,  .949,  .8}85.00 & \cellcolor[rgb]{ 1,  .949,  .8}83.75 & \cellcolor[rgb]{ 1,  .949,  .8}84.89 & \cellcolor[rgb]{ 1,  .949,  .8}84.58 & \cellcolor[rgb]{ .886,  .937,  .855}85.12 & \cellcolor[rgb]{ .886,  .937,  .855}79.72 & \cellcolor[rgb]{ .776,  .878,  .706}85.12 & \cellcolor[rgb]{ .776,  .878,  .706}84.63 & \cellcolor[rgb]{ .776,  .878,  .706}84.86 \\
    \midrule
    \textbf{Bald} & \cellcolor[rgb]{ .988,  .894,  .839}97.72 & \cellcolor[rgb]{ 1,  .949,  .8}98.00 & \cellcolor[rgb]{ 1,  .949,  .8}98.77 & \cellcolor[rgb]{ 1,  .949,  .8}98.90 & \cellcolor[rgb]{ 1,  .949,  .8}99.00 & \cellcolor[rgb]{ 1,  .949,  .8}99.06 & \cellcolor[rgb]{ 1,  .949,  .8}97.82 & \cellcolor[rgb]{ 1,  .949,  .8}98.93 & \cellcolor[rgb]{ .886,  .937,  .855}98.46 & \cellcolor[rgb]{ .886,  .937,  .855}97.88 & \cellcolor[rgb]{ .776,  .878,  .706}98.46 & \cellcolor[rgb]{ .776,  .878,  .706}97.98 & \cellcolor[rgb]{ .776,  .878,  .706}98.03 \\
    \midrule
    \textbf{Bangs} & \cellcolor[rgb]{ .988,  .894,  .839}84.83 & \cellcolor[rgb]{ 1,  .949,  .8}95.00 & \cellcolor[rgb]{ 1,  .949,  .8}95.80 & \cellcolor[rgb]{ 1,  .949,  .8}96.05 & \cellcolor[rgb]{ 1,  .949,  .8}99.00 & \cellcolor[rgb]{ 1,  .949,  .8}96.05 & \cellcolor[rgb]{ 1,  .949,  .8}95.49 & \cellcolor[rgb]{ 1,  .949,  .8}95.93 & \cellcolor[rgb]{ .886,  .937,  .855}95.65 & \cellcolor[rgb]{ .886,  .937,  .855}95.72 & \cellcolor[rgb]{ .776,  .878,  .706}95.72 & \cellcolor[rgb]{ .776,  .878,  .706}95.73 & \cellcolor[rgb]{ .776,  .878,  .706}95.71 \\
    \midrule
    \textbf{Big\_Lips} & \cellcolor[rgb]{ .988,  .894,  .839}75.91 & \cellcolor[rgb]{ 1,  .949,  .8}68.00 & \cellcolor[rgb]{ 1,  .949,  .8}71.48 & \cellcolor[rgb]{ 1,  .949,  .8}71.47 & \cellcolor[rgb]{ 1,  .949,  .8}96.00 & \cellcolor[rgb]{ 1,  .949,  .8}70.88 & \cellcolor[rgb]{ 1,  .949,  .8}71.42 & \cellcolor[rgb]{ 1,  .949,  .8}71.46 & \cellcolor[rgb]{ .886,  .937,  .855}67.29 & \cellcolor[rgb]{ .886,  .937,  .855}67.29 & \cellcolor[rgb]{ .776,  .878,  .706}67.29 & \cellcolor[rgb]{ .776,  .878,  .706}69.78 & \cellcolor[rgb]{ .776,  .878,  .706}69.28 \\
    \midrule
    \textbf{Big\_Nose} & \cellcolor[rgb]{ .988,  .894,  .839}76.44 & \cellcolor[rgb]{ 1,  .949,  .8}78.00 & \cellcolor[rgb]{ 1,  .949,  .8}84.00 & \cellcolor[rgb]{ 1,  .949,  .8}84.53 & \cellcolor[rgb]{ 1,  .949,  .8}85.00 & \cellcolor[rgb]{ 1,  .949,  .8}83.82 & \cellcolor[rgb]{ 1,  .949,  .8}81.83 & \cellcolor[rgb]{ 1,  .949,  .8}83.63 & \cellcolor[rgb]{ .886,  .937,  .855}83.91 & \cellcolor[rgb]{ .886,  .937,  .855}81.85 & \cellcolor[rgb]{ .776,  .878,  .706}83.36 & \cellcolor[rgb]{ .776,  .878,  .706}81.31 & \cellcolor[rgb]{ .776,  .878,  .706}83.81 \\
    \midrule
    \textbf{Black\_Hair} & \cellcolor[rgb]{ .988,  .894,  .839}76.10 & \cellcolor[rgb]{ 1,  .949,  .8}88.00 & \cellcolor[rgb]{ 1,  .949,  .8}89.40 & \cellcolor[rgb]{ 1,  .949,  .8}89.78 & \cellcolor[rgb]{ 1,  .949,  .8}91.00 & \cellcolor[rgb]{ 1,  .949,  .8}90.32 & \cellcolor[rgb]{ 1,  .949,  .8}85.88 & \cellcolor[rgb]{ 1,  .949,  .8}89.84 & \cellcolor[rgb]{ .886,  .937,  .855}88.88 & \cellcolor[rgb]{ .886,  .937,  .855}72.85 & \cellcolor[rgb]{ .776,  .878,  .706}88.88 & \cellcolor[rgb]{ .776,  .878,  .706}88.82 & \cellcolor[rgb]{ .776,  .878,  .706}89.03 \\
    \midrule
    \textbf{Blond\_Hair} & \cellcolor[rgb]{ .988,  .894,  .839}85.09 & \cellcolor[rgb]{ 1,  .949,  .8}95.00 & \cellcolor[rgb]{ 1,  .949,  .8}95.86 & \cellcolor[rgb]{ 1,  .949,  .8}96.01 & \cellcolor[rgb]{ 1,  .949,  .8}96.00 & \cellcolor[rgb]{ 1,  .949,  .8}96.07 & \cellcolor[rgb]{ 1,  .949,  .8}95.17 & \cellcolor[rgb]{ 1,  .949,  .8}95.85 & \cellcolor[rgb]{ .886,  .937,  .855}95.70 & \cellcolor[rgb]{ .886,  .937,  .855}95.68 & \cellcolor[rgb]{ .776,  .878,  .706}95.70 & \cellcolor[rgb]{ .776,  .878,  .706}95.04 & \cellcolor[rgb]{ .776,  .878,  .706}95.76 \\
    \midrule
    \textbf{Blurry} & \cellcolor[rgb]{ .988,  .894,  .839}94.86 & \cellcolor[rgb]{ 1,  .949,  .8}84.00 & \cellcolor[rgb]{ 1,  .949,  .8}95.67 & \cellcolor[rgb]{ 1,  .949,  .8}96.17 & \cellcolor[rgb]{ 1,  .949,  .8}96.00 & \cellcolor[rgb]{ 1,  .949,  .8}95.50 & \cellcolor[rgb]{ 1,  .949,  .8}94.52 & \cellcolor[rgb]{ 1,  .949,  .8}96.11 & \cellcolor[rgb]{ .886,  .937,  .855}95.87 & \cellcolor[rgb]{ .886,  .937,  .855}94.95 & \cellcolor[rgb]{ .776,  .878,  .706}95.87 & \cellcolor[rgb]{ .776,  .878,  .706}95.04 & \cellcolor[rgb]{ .776,  .878,  .706}95.96 \\
    \midrule
    \textbf{Brown\_Hair} & \cellcolor[rgb]{ .988,  .894,  .839}79.61 & \cellcolor[rgb]{ 1,  .949,  .8}80.00 & \cellcolor[rgb]{ 1,  .949,  .8}89.38 & \cellcolor[rgb]{ 1,  .949,  .8}89.15 & \cellcolor[rgb]{ 1,  .949,  .8}88.00 & \cellcolor[rgb]{ 1,  .949,  .8}89.16 & \cellcolor[rgb]{ 1,  .949,  .8}87.72 & \cellcolor[rgb]{ 1,  .949,  .8}88.50 & \cellcolor[rgb]{ .886,  .937,  .855}88.42 & \cellcolor[rgb]{ .886,  .937,  .855}87.64 & \cellcolor[rgb]{ .776,  .878,  .706}88.42 & \cellcolor[rgb]{ .776,  .878,  .706}85.59 & \cellcolor[rgb]{ .776,  .878,  .706}88.25 \\
    \midrule
    \textbf{Bushy\_Eyebrows} & \cellcolor[rgb]{ .988,  .894,  .839}85.63 & \cellcolor[rgb]{ 1,  .949,  .8}90.00 & \cellcolor[rgb]{ 1,  .949,  .8}92.62 & \cellcolor[rgb]{ 1,  .949,  .8}92.84 & \cellcolor[rgb]{ 1,  .949,  .8}92.00 & \cellcolor[rgb]{ 1,  .949,  .8}92.41 & \cellcolor[rgb]{ 1,  .949,  .8}90.59 & \cellcolor[rgb]{ 1,  .949,  .8}92.62 & \cellcolor[rgb]{ .886,  .937,  .855}92.41 & \cellcolor[rgb]{ .886,  .937,  .855}92.20 & \cellcolor[rgb]{ .776,  .878,  .706}92.41 & \cellcolor[rgb]{ .776,  .878,  .706}91.82 & \cellcolor[rgb]{ .776,  .878,  .706}92.66 \\
    \midrule
    \textbf{Chubby} & \cellcolor[rgb]{ .988,  .894,  .839}94.23 & \cellcolor[rgb]{ 1,  .949,  .8}91.00 & \cellcolor[rgb]{ 1,  .949,  .8}95.44 & \cellcolor[rgb]{ 1,  .949,  .8}95.67 & \cellcolor[rgb]{ 1,  .949,  .8}96.00 & \cellcolor[rgb]{ 1,  .949,  .8}94.98 & \cellcolor[rgb]{ 1,  .949,  .8}95.10 & \cellcolor[rgb]{ 1,  .949,  .8}95.46 & \cellcolor[rgb]{ .886,  .937,  .855}94.69 & \cellcolor[rgb]{ .886,  .937,  .855}94.69 & \cellcolor[rgb]{ .776,  .878,  .706}94.69 & \cellcolor[rgb]{ .776,  .878,  .706}93.90 & \cellcolor[rgb]{ .776,  .878,  .706}94.94 \\
    \midrule
    \textbf{Double\_Chin} & \cellcolor[rgb]{ .988,  .894,  .839}95.35 & \cellcolor[rgb]{ 1,  .949,  .8}92.00 & \cellcolor[rgb]{ 1,  .949,  .8}96.32 & \cellcolor[rgb]{ 1,  .949,  .8}96.32 & \cellcolor[rgb]{ 1,  .949,  .8}97.00 & \cellcolor[rgb]{ 1,  .949,  .8}96.18 & \cellcolor[rgb]{ 1,  .949,  .8}95.94 & \cellcolor[rgb]{ 1,  .949,  .8}96.26 & \cellcolor[rgb]{ .886,  .937,  .855}95.43 & \cellcolor[rgb]{ .886,  .937,  .855}95.43 & \cellcolor[rgb]{ .776,  .878,  .706}95.68 & \cellcolor[rgb]{ .776,  .878,  .706}95.23 & \cellcolor[rgb]{ .776,  .878,  .706}95.80 \\
    \midrule
    \textbf{Eyeglasses} & \cellcolor[rgb]{ .988,  .894,  .839}93.54 & \cellcolor[rgb]{ 1,  .949,  .8}99.00 & \cellcolor[rgb]{ 1,  .949,  .8}99.47 & \cellcolor[rgb]{ 1,  .949,  .8}99.63 & \cellcolor[rgb]{ 1,  .949,  .8}99.00 & \cellcolor[rgb]{ 1,  .949,  .8}99.61 & \cellcolor[rgb]{ 1,  .949,  .8}99.38 & \cellcolor[rgb]{ 1,  .949,  .8}99.59 & \cellcolor[rgb]{ .886,  .937,  .855}99.43 & \cellcolor[rgb]{ .886,  .937,  .855}99.48 & \cellcolor[rgb]{ .776,  .878,  .706}99.30 & \cellcolor[rgb]{ .776,  .878,  .706}99.58 & \cellcolor[rgb]{ .776,  .878,  .706}99.51 \\
    \midrule
    \textbf{Goatee} & \cellcolor[rgb]{ .988,  .894,  .839}93.65 & \cellcolor[rgb]{ 1,  .949,  .8}95.00 & \cellcolor[rgb]{ 1,  .949,  .8}97.04 & \cellcolor[rgb]{ 1,  .949,  .8}97.24 & \cellcolor[rgb]{ 1,  .949,  .8}99.00 & \cellcolor[rgb]{ 1,  .949,  .8}97.31 & \cellcolor[rgb]{ 1,  .949,  .8}97.21 & \cellcolor[rgb]{ 1,  .949,  .8}97.38 & \cellcolor[rgb]{ .886,  .937,  .855}96.51 & \cellcolor[rgb]{ .886,  .937,  .855}95.41 & \cellcolor[rgb]{ .776,  .878,  .706}96.70 & \cellcolor[rgb]{ .776,  .878,  .706}95.88 & \cellcolor[rgb]{ .776,  .878,  .706}96.68 \\
    \midrule
    \textbf{Gray\_Hair} & \cellcolor[rgb]{ .988,  .894,  .839}95.76 & \cellcolor[rgb]{ 1,  .949,  .8}97.00 & \cellcolor[rgb]{ 1,  .949,  .8}98.10 & \cellcolor[rgb]{ 1,  .949,  .8}98.20 & \cellcolor[rgb]{ 1,  .949,  .8}98.00 & \cellcolor[rgb]{ 1,  .949,  .8}98.28 & \cellcolor[rgb]{ 1,  .949,  .8}97.89 & \cellcolor[rgb]{ 1,  .949,  .8}98.21 & \cellcolor[rgb]{ .886,  .937,  .855}97.57 & \cellcolor[rgb]{ .886,  .937,  .855}95.99 & \cellcolor[rgb]{ .776,  .878,  .706}97.57 & \cellcolor[rgb]{ .776,  .878,  .706}95.80 & \cellcolor[rgb]{ .776,  .878,  .706}97.45 \\
    \midrule
    \textbf{Heavy\_Makeup} & \cellcolor[rgb]{ .988,  .894,  .839}61.57 & \cellcolor[rgb]{ 1,  .949,  .8}90.00 & \cellcolor[rgb]{ 1,  .949,  .8}90.99 & \cellcolor[rgb]{ 1,  .949,  .8}91.55 & \cellcolor[rgb]{ 1,  .949,  .8}92.00 & \cellcolor[rgb]{ 1,  .949,  .8}91.10 & \cellcolor[rgb]{ 1,  .949,  .8}90.82 & \cellcolor[rgb]{ 1,  .949,  .8}91.53 & \cellcolor[rgb]{ .886,  .937,  .855}91.18 & \cellcolor[rgb]{ .886,  .937,  .855}91.51 & \cellcolor[rgb]{ .776,  .878,  .706}91.51 & \cellcolor[rgb]{ .776,  .878,  .706}91.55 & \cellcolor[rgb]{ .776,  .878,  .706}91.59 \\
    \midrule
    \textbf{High\_Cheekbones} & \cellcolor[rgb]{ .988,  .894,  .839}54.76 & \cellcolor[rgb]{ 1,  .949,  .8}87.00 & \cellcolor[rgb]{ 1,  .949,  .8}87.01 & \cellcolor[rgb]{ 1,  .949,  .8}87.58 & \cellcolor[rgb]{ 1,  .949,  .8}88.00 & \cellcolor[rgb]{ 1,  .949,  .8}86.88 & \cellcolor[rgb]{ 1,  .949,  .8}86.11 & \cellcolor[rgb]{ 1,  .949,  .8}87.44 & \cellcolor[rgb]{ .886,  .937,  .855}87.08 & \cellcolor[rgb]{ .886,  .937,  .855}87.54 & \cellcolor[rgb]{ .776,  .878,  .706}87.54 & \cellcolor[rgb]{ .776,  .878,  .706}87.62 & \cellcolor[rgb]{ .776,  .878,  .706}87.61 \\
    \midrule
    \textbf{Male} & \cellcolor[rgb]{ .988,  .894,  .839}58.06 & \cellcolor[rgb]{ 1,  .949,  .8}98.00 & \cellcolor[rgb]{ 1,  .949,  .8}98.10 & \cellcolor[rgb]{ 1,  .949,  .8}98.17 & \cellcolor[rgb]{ 1,  .949,  .8}98.00 & \cellcolor[rgb]{ 1,  .949,  .8}98.26 & \cellcolor[rgb]{ 1,  .949,  .8}97.29 & \cellcolor[rgb]{ 1,  .949,  .8}98.39 & \cellcolor[rgb]{ .886,  .937,  .855}97.58 & \cellcolor[rgb]{ .886,  .937,  .855}98.14 & \cellcolor[rgb]{ .776,  .878,  .706}98.14 & \cellcolor[rgb]{ .776,  .878,  .706}98.09 & \cellcolor[rgb]{ .776,  .878,  .706}97.95 \\
    \midrule
    \textbf{Mouth\_Slightly\_Open} & \cellcolor[rgb]{ .988,  .894,  .839}51.78 & \cellcolor[rgb]{ 1,  .949,  .8}92.00 & \cellcolor[rgb]{ 1,  .949,  .8}93.54 & \cellcolor[rgb]{ 1,  .949,  .8}93.74 & \cellcolor[rgb]{ 1,  .949,  .8}94.00 & \cellcolor[rgb]{ 1,  .949,  .8}92.60 & \cellcolor[rgb]{ 1,  .949,  .8}92.82 & \cellcolor[rgb]{ 1,  .949,  .8}94.05 & \cellcolor[rgb]{ .886,  .937,  .855}93.62 & \cellcolor[rgb]{ .886,  .937,  .855}93.91 & \cellcolor[rgb]{ .776,  .878,  .706}93.91 & \cellcolor[rgb]{ .776,  .878,  .706}93.90 & \cellcolor[rgb]{ .776,  .878,  .706}93.78 \\
    \midrule
    \textbf{Mustache} & \cellcolor[rgb]{ .988,  .894,  .839}95.92 & \cellcolor[rgb]{ 1,  .949,  .8}95.00 & \cellcolor[rgb]{ 1,  .949,  .8}96.82 & \cellcolor[rgb]{ 1,  .949,  .8}96.88 & \cellcolor[rgb]{ 1,  .949,  .8}97.00 & \cellcolor[rgb]{ 1,  .949,  .8}96.89 & \cellcolor[rgb]{ 1,  .949,  .8}96.89 & \cellcolor[rgb]{ 1,  .949,  .8}96.90 & \cellcolor[rgb]{ .886,  .937,  .855}96.12 & \cellcolor[rgb]{ .886,  .937,  .855}96.12 & \cellcolor[rgb]{ .776,  .878,  .706}96.12 & \cellcolor[rgb]{ .776,  .878,  .706}96.16 & \cellcolor[rgb]{ .776,  .878,  .706}95.86 \\
    \midrule
    \textbf{Narrow\_Eyes} & \cellcolor[rgb]{ .988,  .894,  .839}88.41 & \cellcolor[rgb]{ 1,  .949,  .8}81.00 & \cellcolor[rgb]{ 1,  .949,  .8}86.52 & \cellcolor[rgb]{ 1,  .949,  .8}87.23 & \cellcolor[rgb]{ 1,  .949,  .8}90.00 & \cellcolor[rgb]{ 1,  .949,  .8}87.23 & \cellcolor[rgb]{ 1,  .949,  .8}87.15 & \cellcolor[rgb]{ 1,  .949,  .8}87.56 & \cellcolor[rgb]{ .886,  .937,  .855}86.79 & \cellcolor[rgb]{ .886,  .937,  .855}85.13 & \cellcolor[rgb]{ .776,  .878,  .706}86.84 & \cellcolor[rgb]{ .776,  .878,  .706}87.31 & \cellcolor[rgb]{ .776,  .878,  .706}86.88 \\
    \midrule
    \textbf{No\_Beard} & \cellcolor[rgb]{ .988,  .894,  .839}83.42 & \cellcolor[rgb]{ 1,  .949,  .8}95.00 & \cellcolor[rgb]{ 1,  .949,  .8}95.58 & \cellcolor[rgb]{ 1,  .949,  .8}96.05 & \cellcolor[rgb]{ 1,  .949,  .8}97.00 & \cellcolor[rgb]{ 1,  .949,  .8}95.99 & \cellcolor[rgb]{ 1,  .949,  .8}95.33 & \cellcolor[rgb]{ 1,  .949,  .8}96.22 & \cellcolor[rgb]{ .886,  .937,  .855}95.77 & \cellcolor[rgb]{ .886,  .937,  .855}96.17 & \cellcolor[rgb]{ .776,  .878,  .706}96.17 & \cellcolor[rgb]{ .776,  .878,  .706}95.57 & \cellcolor[rgb]{ .776,  .878,  .706}96.17 \\
    \midrule
    \textbf{Oval\_Face} & \cellcolor[rgb]{ .988,  .894,  .839}71.68 & \cellcolor[rgb]{ 1,  .949,  .8}66.00 & \cellcolor[rgb]{ 1,  .949,  .8}75.73 & \cellcolor[rgb]{ 1,  .949,  .8}75.84 & \cellcolor[rgb]{ 1,  .949,  .8}78.00 & \cellcolor[rgb]{ 1,  .949,  .8}75.79 & \cellcolor[rgb]{ 1,  .949,  .8}74.87 & \cellcolor[rgb]{ 1,  .949,  .8}75.03 & \cellcolor[rgb]{ .886,  .937,  .855}75.40 & \cellcolor[rgb]{ .886,  .937,  .855}70.45 & \cellcolor[rgb]{ .776,  .878,  .706}75.40 & \cellcolor[rgb]{ .776,  .878,  .706}75.75 & \cellcolor[rgb]{ .776,  .878,  .706}74.93 \\
    \midrule
    \textbf{Pale\_Skin} & \cellcolor[rgb]{ .988,  .894,  .839}95.70 & \cellcolor[rgb]{ 1,  .949,  .8}91.00 & \cellcolor[rgb]{ 1,  .949,  .8}97.00 & \cellcolor[rgb]{ 1,  .949,  .8}97.05 & \cellcolor[rgb]{ 1,  .949,  .8}97.00 & \cellcolor[rgb]{ 1,  .949,  .8}97.04 & \cellcolor[rgb]{ 1,  .949,  .8}96.97 & \cellcolor[rgb]{ 1,  .949,  .8}97.08 & \cellcolor[rgb]{ .886,  .937,  .855}96.90 & \cellcolor[rgb]{ .886,  .937,  .855}95.80 & \cellcolor[rgb]{ .776,  .878,  .706}96.90 & \cellcolor[rgb]{ .776,  .878,  .706}96.72 & \cellcolor[rgb]{ .776,  .878,  .706}97.00 \\
    \midrule
    \textbf{Pointy\_Nose} & \cellcolor[rgb]{ .988,  .894,  .839}72.45 & \cellcolor[rgb]{ 1,  .949,  .8}72.00 & \cellcolor[rgb]{ 1,  .949,  .8}76.46 & \cellcolor[rgb]{ 1,  .949,  .8}77.47 & \cellcolor[rgb]{ 1,  .949,  .8}78.00 & \cellcolor[rgb]{ 1,  .949,  .8}74.83 & \cellcolor[rgb]{ 1,  .949,  .8}76.24 & \cellcolor[rgb]{ 1,  .949,  .8}77.35 & \cellcolor[rgb]{ .886,  .937,  .855}76.13 & \cellcolor[rgb]{ .886,  .937,  .855}71.45 & \cellcolor[rgb]{ .776,  .878,  .706}76.13 & \cellcolor[rgb]{ .776,  .878,  .706}76.46 & \cellcolor[rgb]{ .776,  .878,  .706}76.47 \\
    \midrule
    \textbf{Receding\_Hairline} & \cellcolor[rgb]{ .988,  .894,  .839}91.99 & \cellcolor[rgb]{ 1,  .949,  .8}89.00 & \cellcolor[rgb]{ 1,  .949,  .8}93.56 & \cellcolor[rgb]{ 1,  .949,  .8}93.81 & \cellcolor[rgb]{ 1,  .949,  .8}94.00 & \cellcolor[rgb]{ 1,  .949,  .8}93.29 & \cellcolor[rgb]{ 1,  .949,  .8}91.74 & \cellcolor[rgb]{ 1,  .949,  .8}93.44 & \cellcolor[rgb]{ .886,  .937,  .855}92.55 & \cellcolor[rgb]{ .886,  .937,  .855}91.52 & \cellcolor[rgb]{ .776,  .878,  .706}92.55 & \cellcolor[rgb]{ .776,  .878,  .706}92.40 & \cellcolor[rgb]{ .776,  .878,  .706}92.25 \\
    \midrule
    \textbf{Rosy\_Cheeks} & \cellcolor[rgb]{ .988,  .894,  .839}93.53 & \cellcolor[rgb]{ 1,  .949,  .8}90.00 & \cellcolor[rgb]{ 1,  .949,  .8}94.82 & \cellcolor[rgb]{ 1,  .949,  .8}95.16 & \cellcolor[rgb]{ 1,  .949,  .8}96.00 & \cellcolor[rgb]{ 1,  .949,  .8}94.45 & \cellcolor[rgb]{ 1,  .949,  .8}94.54 & \cellcolor[rgb]{ 1,  .949,  .8}95.07 & \cellcolor[rgb]{ .886,  .937,  .855}94.59 & \cellcolor[rgb]{ .886,  .937,  .855}92.83 & \cellcolor[rgb]{ .776,  .878,  .706}94.59 & \cellcolor[rgb]{ .776,  .878,  .706}94.51 & \cellcolor[rgb]{ .776,  .878,  .706}94.79 \\
    \midrule
    \textbf{Sideburns} & \cellcolor[rgb]{ .988,  .894,  .839}94.37 & \cellcolor[rgb]{ 1,  .949,  .8}96.00 & \cellcolor[rgb]{ 1,  .949,  .8}97.59 & \cellcolor[rgb]{ 1,  .949,  .8}97.85 & \cellcolor[rgb]{ 1,  .949,  .8}98.00 & \cellcolor[rgb]{ 1,  .949,  .8}97.83 & \cellcolor[rgb]{ 1,  .949,  .8}97.46 & \cellcolor[rgb]{ 1,  .949,  .8}97.64 & \cellcolor[rgb]{ .886,  .937,  .855}96.83 & \cellcolor[rgb]{ .886,  .937,  .855}96.09 & \cellcolor[rgb]{ .776,  .878,  .706}96.83 & \cellcolor[rgb]{ .776,  .878,  .706}96.01 & \cellcolor[rgb]{ .776,  .878,  .706}97.17 \\
    \midrule
    \textbf{Smiling} & \cellcolor[rgb]{ .988,  .894,  .839}52.03 & \cellcolor[rgb]{ 1,  .949,  .8}92.00 & \cellcolor[rgb]{ 1,  .949,  .8}92.60 & \cellcolor[rgb]{ 1,  .949,  .8}92.73 & \cellcolor[rgb]{ 1,  .949,  .8}94.00 & \cellcolor[rgb]{ 1,  .949,  .8}91.77 & \cellcolor[rgb]{ 1,  .949,  .8}90.45 & \cellcolor[rgb]{ 1,  .949,  .8}92.73 & \cellcolor[rgb]{ .886,  .937,  .855}92.42 & \cellcolor[rgb]{ .886,  .937,  .855}92.74 & \cellcolor[rgb]{ .776,  .878,  .706}92.74 & \cellcolor[rgb]{ .776,  .878,  .706}92.89 & \cellcolor[rgb]{ .776,  .878,  .706}92.70 \\
    \midrule
    \textbf{Straight\_Hair} & \cellcolor[rgb]{ .988,  .894,  .839}79.14 & \cellcolor[rgb]{ 1,  .949,  .8}73.00 & \cellcolor[rgb]{ 1,  .949,  .8}82.26 & \cellcolor[rgb]{ 1,  .949,  .8}83.58 & \cellcolor[rgb]{ 1,  .949,  .8}85.00 & \cellcolor[rgb]{ 1,  .949,  .8}84.10 & \cellcolor[rgb]{ 1,  .949,  .8}82.17 & \cellcolor[rgb]{ 1,  .949,  .8}83.52 & \cellcolor[rgb]{ .886,  .937,  .855}83.11 & \cellcolor[rgb]{ .886,  .937,  .855}79.04 & \cellcolor[rgb]{ .776,  .878,  .706}83.11 & \cellcolor[rgb]{ .776,  .878,  .706}82.36 & \cellcolor[rgb]{ .776,  .878,  .706}80.41 \\
    \midrule
    \textbf{Wavy\_Hair} & \cellcolor[rgb]{ .988,  .894,  .839}68.06 & \cellcolor[rgb]{ 1,  .949,  .8}80.00 & \cellcolor[rgb]{ 1,  .949,  .8}82.47 & \cellcolor[rgb]{ 1,  .949,  .8}83.91 & \cellcolor[rgb]{ 1,  .949,  .8}87.00 & \cellcolor[rgb]{ 1,  .949,  .8}85.65 & \cellcolor[rgb]{ 1,  .949,  .8}83.37 & \cellcolor[rgb]{ 1,  .949,  .8}84.07 & \cellcolor[rgb]{ .886,  .937,  .855}83.28 & \cellcolor[rgb]{ .886,  .937,  .855}63.58 & \cellcolor[rgb]{ .776,  .878,  .706}83.28 & \cellcolor[rgb]{ .776,  .878,  .706}83.10 & \cellcolor[rgb]{ .776,  .878,  .706}81.70 \\
    \midrule
    \textbf{Wearing\_Earrings} & \cellcolor[rgb]{ .988,  .894,  .839}81.35 & \cellcolor[rgb]{ 1,  .949,  .8}82.00 & \cellcolor[rgb]{ 1,  .949,  .8}89.60 & \cellcolor[rgb]{ 1,  .949,  .8}90.43 & \cellcolor[rgb]{ 1,  .949,  .8}91.00 & \cellcolor[rgb]{ 1,  .949,  .8}90.20 & \cellcolor[rgb]{ 1,  .949,  .8}90.33 & \cellcolor[rgb]{ 1,  .949,  .8}89.93 & \cellcolor[rgb]{ .886,  .937,  .855}90.41 & \cellcolor[rgb]{ .886,  .937,  .855}90.48 & \cellcolor[rgb]{ .776,  .878,  .706}90.41 & \cellcolor[rgb]{ .776,  .878,  .706}89.72 & \cellcolor[rgb]{ .776,  .878,  .706}89.44 \\
    \midrule
    \textbf{Wearing\_Hat} & \cellcolor[rgb]{ .988,  .894,  .839}95.06 & \cellcolor[rgb]{ 1,  .949,  .8}99.00 & \cellcolor[rgb]{ 1,  .949,  .8}98.95 & \cellcolor[rgb]{ 1,  .949,  .8}99.05 & \cellcolor[rgb]{ 1,  .949,  .8}99.00 & \cellcolor[rgb]{ 1,  .949,  .8}99.02 & \cellcolor[rgb]{ 1,  .949,  .8}98.66 & \cellcolor[rgb]{ 1,  .949,  .8}99.02 & \cellcolor[rgb]{ .886,  .937,  .855}98.71 & \cellcolor[rgb]{ .886,  .937,  .855}95.79 & \cellcolor[rgb]{ .776,  .878,  .706}98.71 & \cellcolor[rgb]{ .776,  .878,  .706}98.42 & \cellcolor[rgb]{ .776,  .878,  .706}98.74 \\
    \midrule
    \textbf{Wearing\_Lipstick} & \cellcolor[rgb]{ .988,  .894,  .839}53.04 & \cellcolor[rgb]{ 1,  .949,  .8}93.00 & \cellcolor[rgb]{ 1,  .949,  .8}93.93 & \cellcolor[rgb]{ 1,  .949,  .8}94.11 & \cellcolor[rgb]{ 1,  .949,  .8}93.00 & \cellcolor[rgb]{ 1,  .949,  .8}91.69 & \cellcolor[rgb]{ 1,  .949,  .8}92.99 & \cellcolor[rgb]{ 1,  .949,  .8}94.24 & \cellcolor[rgb]{ .886,  .937,  .855}92.66 & \cellcolor[rgb]{ .886,  .937,  .855}93.23 & \cellcolor[rgb]{ .776,  .878,  .706}93.23 & \cellcolor[rgb]{ .776,  .878,  .706}94.00 & \cellcolor[rgb]{ .776,  .878,  .706}93.21 \\
    \midrule
    \textbf{Wearing\_Necklace} & \cellcolor[rgb]{ .988,  .894,  .839}87.86 & \cellcolor[rgb]{ 1,  .949,  .8}71.00 & \cellcolor[rgb]{ 1,  .949,  .8}87.04 & \cellcolor[rgb]{ 1,  .949,  .8}86.63 & \cellcolor[rgb]{ 1,  .949,  .8}89.00 & \cellcolor[rgb]{ 1,  .949,  .8}87.85 & \cellcolor[rgb]{ 1,  .949,  .8}87.55 & \cellcolor[rgb]{ 1,  .949,  .8}87.70 & \cellcolor[rgb]{ .886,  .937,  .855}87.54 & \cellcolor[rgb]{ .886,  .937,  .855}86.22 & \cellcolor[rgb]{ .776,  .878,  .706}87.54 & \cellcolor[rgb]{ .776,  .878,  .706}87.50 & \cellcolor[rgb]{ .776,  .878,  .706}85.61 \\
    \midrule
    \textbf{Wearing\_Necktie} & \cellcolor[rgb]{ .988,  .894,  .839}92.70 & \cellcolor[rgb]{ 1,  .949,  .8}93.00 & \cellcolor[rgb]{ 1,  .949,  .8}96.63 & \cellcolor[rgb]{ 1,  .949,  .8}96.51 & \cellcolor[rgb]{ 1,  .949,  .8}97.00 & \cellcolor[rgb]{ 1,  .949,  .8}96.90 & \cellcolor[rgb]{ 1,  .949,  .8}96.43 & \cellcolor[rgb]{ 1,  .949,  .8}96.85 & \cellcolor[rgb]{ .886,  .937,  .855}96.66 & \cellcolor[rgb]{ .886,  .937,  .855}95.61 & \cellcolor[rgb]{ .776,  .878,  .706}96.66 & \cellcolor[rgb]{ .776,  .878,  .706}95.24 & \cellcolor[rgb]{ .776,  .878,  .706}96.05 \\
    \midrule
    \textbf{Young} & \cellcolor[rgb]{ .988,  .894,  .839}77.89 & \cellcolor[rgb]{ 1,  .949,  .8}87.00 & \cellcolor[rgb]{ 1,  .949,  .8}88.08 & \cellcolor[rgb]{ 1,  .949,  .8}88.48 & \cellcolor[rgb]{ 1,  .949,  .8}90.00 & \cellcolor[rgb]{ 1,  .949,  .8}88.66 & \cellcolor[rgb]{ 1,  .949,  .8}86.21 & \cellcolor[rgb]{ 1,  .949,  .8}88.59 & \cellcolor[rgb]{ .886,  .937,  .855}87.95 & \cellcolor[rgb]{ .886,  .937,  .855}88.45 & \cellcolor[rgb]{ .776,  .878,  .706}88.45 & \cellcolor[rgb]{ .776,  .878,  .706}86.93 & \cellcolor[rgb]{ .776,  .878,  .706}88.01 \\
    \midrule
    \textbf{Mean Accuracy} & \cellcolor[rgb]{ .988,  .894,  .839}\textbf{80.57} & \cellcolor[rgb]{ 1,  .949,  .8}\textbf{87.30} & \cellcolor[rgb]{ 1,  .949,  .8}\textbf{90.94} & \cellcolor[rgb]{ 1,  .949,  .8}\textbf{91.29} & \cellcolor[rgb]{ 1,  .949,  .8}\textbf{92.60} & \cellcolor[rgb]{ 1,  .949,  .8}\textbf{91.01} & \cellcolor[rgb]{ 1,  .949,  .8}\textbf{90.32} & \cellcolor[rgb]{ 1,  .949,  .8}\textbf{91.23} & \cellcolor[rgb]{ .886,  .937,  .855}\textbf{90.72} & \cellcolor[rgb]{ .886,  .937,  .855}\textbf{88.87} & \cellcolor[rgb]{ .886,  .937,  .855}\textbf{90.80} & \cellcolor[rgb]{ .776,  .878,  .706}\textbf{90.42} & \cellcolor[rgb]{ .776,  .878,  .706}\textbf{90.61} \\
    \end{tabular}%
  \label{tab:PerformanceCompareOnCelebA}%
  	}
\end{table*}%

\begin{table*}
  \centering
  \caption{Attribute detection performance comparison on the LFWA dataset in terms of individual and mean detection accuracy for the attributes.}
  \scalebox{0.7}{
    \begin{tabular}{p{10.59em}ccccccccc}
    \toprule
     \multirow{3}[6]{*}{\textbf{Attributes}} & \cellcolor[rgb]{ .988,  .894,  .839} & \cellcolor[rgb]{ 1,  .949,  .8} & \cellcolor[rgb]{ 1,  .949,  .8} & \cellcolor[rgb]{ 1,  .949,  .8} & \cellcolor[rgb]{ .886,  .937,  .855}\textbf{Proposed} & \cellcolor[rgb]{ .886,  .937,  .855}\textbf{Proposed} & \multicolumn{3}{c}{\cellcolor[rgb]{ .776,  .878,  .706}\textbf{Proposed Committee Machine }} \\
\cmidrule{8-10}    \multicolumn{1}{c}{} & \multicolumn{1}{c}{\cellcolor[rgb]{ .988,  .894,  .839}\textbf{Prior}} & \multicolumn{1}{c}{\cellcolor[rgb]{ 1,  .949,  .8}\textbf{LENet+}} & \multicolumn{1}{c}{\cellcolor[rgb]{ 1,  .949,  .8}\textbf{MCNN+}} & \multicolumn{1}{p{5.32em}}{\cellcolor[rgb]{ 1,  .949,  .8}\textbf{DMTL\cite{HuHan_DeepMultiTaskHeteroAttrib}}} & \multicolumn{1}{c}{\cellcolor[rgb]{ .886,  .937,  .855}\textbf{FULL}} & \multicolumn{1}{c}{\cellcolor[rgb]{ .886,  .937,  .855}\textbf{GP}} & \multicolumn{1}{p{6.09em}}{\cellcolor[rgb]{ .776,  .878,  .706}\textbf{ HRP  }} & \multicolumn{2}{c}{\cellcolor[rgb]{ .776,  .878,  .706}\textbf{NSA}} \\
\cmidrule{9-10}    \multicolumn{1}{c}{} & \cellcolor[rgb]{ .988,  .894,  .839} & \multicolumn{1}{c}{\cellcolor[rgb]{ 1,  .949,  .8}\textbf{Anet\cite{CelebA_liu2015faceattributes}}} & \multicolumn{1}{c}{\cellcolor[rgb]{ 1,  .949,  .8}\textbf{AUX\cite{EmilyNet_AAI_Multitask}}} & \cellcolor[rgb]{ 1,  .949,  .8} & \cellcolor[rgb]{ .886,  .937,  .855} & \cellcolor[rgb]{ .886,  .937,  .855} & \multicolumn{1}{c}{\cellcolor[rgb]{ .776,  .878,  .706}\textbf{ }} & \multicolumn{1}{c}{\cellcolor[rgb]{ .776,  .878,  .706}\textbf{Prod. Rule}} & \multicolumn{1}{c}{\cellcolor[rgb]{ .776,  .878,  .706}\textbf{Med. Rule}} \\
    \midrule
    \textbf{5\_o\_Clock\_Shadow} & \cellcolor[rgb]{ .988,  .894,  .839}59.76 & \cellcolor[rgb]{ 1,  .949,  .8}84 & \cellcolor[rgb]{ 1,  .949,  .8}77.06 & \cellcolor[rgb]{ 1,  .949,  .8}80 & \cellcolor[rgb]{ .886,  .937,  .855}74.72 & \cellcolor[rgb]{ .886,  .937,  .855}74.72 & \cellcolor[rgb]{ .776,  .878,  .706}74.72 & \cellcolor[rgb]{ .776,  .878,  .706}77.47 & \cellcolor[rgb]{ .776,  .878,  .706}77.59 \\
    \midrule
    \textbf{Arched\_Eyebrows} & \cellcolor[rgb]{ .988,  .894,  .839}72.35 & \cellcolor[rgb]{ 1,  .949,  .8}82 & \cellcolor[rgb]{ 1,  .949,  .8}81.78 & \cellcolor[rgb]{ 1,  .949,  .8}86 & \cellcolor[rgb]{ .886,  .937,  .855}78.78 & \cellcolor[rgb]{ .886,  .937,  .855}78.78 & \cellcolor[rgb]{ .776,  .878,  .706}78.78 & \cellcolor[rgb]{ .776,  .878,  .706}81.82 & \cellcolor[rgb]{ .776,  .878,  .706}81.72 \\
    \midrule
    \textbf{Attractive} & \cellcolor[rgb]{ .988,  .894,  .839}62.09 & \cellcolor[rgb]{ 1,  .949,  .8}83 & \cellcolor[rgb]{ 1,  .949,  .8}80.31 & \cellcolor[rgb]{ 1,  .949,  .8}82 & \cellcolor[rgb]{ .886,  .937,  .855}77.44 & \cellcolor[rgb]{ .886,  .937,  .855}77.44 & \cellcolor[rgb]{ .776,  .878,  .706}77.44 & \cellcolor[rgb]{ .776,  .878,  .706}80.25 & \cellcolor[rgb]{ .776,  .878,  .706}80.16 \\
    \midrule
    \textbf{Bags\_Under\_Eyes} & \cellcolor[rgb]{ .988,  .894,  .839}59.52 & \cellcolor[rgb]{ 1,  .949,  .8}83 & \cellcolor[rgb]{ 1,  .949,  .8}83.48 & \cellcolor[rgb]{ 1,  .949,  .8}84 & \cellcolor[rgb]{ .886,  .937,  .855}79.11 & \cellcolor[rgb]{ .886,  .937,  .855}79.11 & \cellcolor[rgb]{ .776,  .878,  .706}79.11 & \cellcolor[rgb]{ .776,  .878,  .706}82.98 & \cellcolor[rgb]{ .776,  .878,  .706}82.62 \\
    \midrule
    \textbf{Bald} & \cellcolor[rgb]{ .988,  .894,  .839}88.94 & \cellcolor[rgb]{ 1,  .949,  .8}88 & \cellcolor[rgb]{ 1,  .949,  .8}91.94 & \cellcolor[rgb]{ 1,  .949,  .8}92 & \cellcolor[rgb]{ .886,  .937,  .855}91.69 & \cellcolor[rgb]{ .886,  .937,  .855}91.51 & \cellcolor[rgb]{ .776,  .878,  .706}91.51 & \cellcolor[rgb]{ .776,  .878,  .706}90.97 & \cellcolor[rgb]{ .776,  .878,  .706}91.88 \\
    \midrule
    \textbf{Bangs} & \cellcolor[rgb]{ .988,  .894,  .839}83.57 & \cellcolor[rgb]{ 1,  .949,  .8}88 & \cellcolor[rgb]{ 1,  .949,  .8}90.08 & \cellcolor[rgb]{ 1,  .949,  .8}93 & \cellcolor[rgb]{ .886,  .937,  .855}89.72 & \cellcolor[rgb]{ .886,  .937,  .855}89.72 & \cellcolor[rgb]{ .776,  .878,  .706}89.72 & \cellcolor[rgb]{ .776,  .878,  .706}90.89 & \cellcolor[rgb]{ .776,  .878,  .706}90.71 \\
    \midrule
    \textbf{Big\_Lips} & \cellcolor[rgb]{ .988,  .894,  .839}64.07 & \cellcolor[rgb]{ 1,  .949,  .8}75 & \cellcolor[rgb]{ 1,  .949,  .8}79.24 & \cellcolor[rgb]{ 1,  .949,  .8}77 & \cellcolor[rgb]{ .886,  .937,  .855}75.47 & \cellcolor[rgb]{ .886,  .937,  .855}77.54 & \cellcolor[rgb]{ .776,  .878,  .706}77.54 & \cellcolor[rgb]{ .776,  .878,  .706}79.10 & \cellcolor[rgb]{ .776,  .878,  .706}78.97 \\
    \midrule
    \textbf{Big\_Nose} & \cellcolor[rgb]{ .988,  .894,  .839}69.62 & \cellcolor[rgb]{ 1,  .949,  .8}81 & \cellcolor[rgb]{ 1,  .949,  .8}84.98 & \cellcolor[rgb]{ 1,  .949,  .8}83 & \cellcolor[rgb]{ .886,  .937,  .855}80.23 & \cellcolor[rgb]{ .886,  .937,  .855}80.23 & \cellcolor[rgb]{ .776,  .878,  .706}80.23 & \cellcolor[rgb]{ .776,  .878,  .706}82.95 & \cellcolor[rgb]{ .776,  .878,  .706}83.13 \\
    \midrule
    \textbf{Black\_Hair} & \cellcolor[rgb]{ .988,  .894,  .839}85.53 & \cellcolor[rgb]{ 1,  .949,  .8}90 & \cellcolor[rgb]{ 1,  .949,  .8}92.63 & \cellcolor[rgb]{ 1,  .949,  .8}92 & \cellcolor[rgb]{ .886,  .937,  .855}91.63 & \cellcolor[rgb]{ .886,  .937,  .855}92.22 & \cellcolor[rgb]{ .776,  .878,  .706}92.22 & \cellcolor[rgb]{ .776,  .878,  .706}92.34 & \cellcolor[rgb]{ .776,  .878,  .706}92.49 \\
    \midrule
    \textbf{Blond\_Hair} & \cellcolor[rgb]{ .988,  .894,  .839}95.75 & \cellcolor[rgb]{ 1,  .949,  .8}97 & \cellcolor[rgb]{ 1,  .949,  .8}97.41 & \cellcolor[rgb]{ 1,  .949,  .8}97 & \cellcolor[rgb]{ .886,  .937,  .855}97.31 & \cellcolor[rgb]{ .886,  .937,  .855}97.31 & \cellcolor[rgb]{ .776,  .878,  .706}97.31 & \cellcolor[rgb]{ .776,  .878,  .706}97.47 & \cellcolor[rgb]{ .776,  .878,  .706}97.47 \\
    \midrule
    \textbf{Blurry} & \cellcolor[rgb]{ .988,  .894,  .839}84.66 & \cellcolor[rgb]{ 1,  .949,  .8}74 & \cellcolor[rgb]{ 1,  .949,  .8}85.23 & \cellcolor[rgb]{ 1,  .949,  .8}89 & \cellcolor[rgb]{ .886,  .937,  .855}85.41 & \cellcolor[rgb]{ .886,  .937,  .855}85.41 & \cellcolor[rgb]{ .776,  .878,  .706}85.41 & \cellcolor[rgb]{ .776,  .878,  .706}86.41 & \cellcolor[rgb]{ .776,  .878,  .706}86.42 \\
    \midrule
    \textbf{Brown\_Hair} & \cellcolor[rgb]{ .988,  .894,  .839}62.02 & \cellcolor[rgb]{ 1,  .949,  .8}77 & \cellcolor[rgb]{ 1,  .949,  .8}80.85 & \cellcolor[rgb]{ 1,  .949,  .8}81 & \cellcolor[rgb]{ .886,  .937,  .855}79.22 & \cellcolor[rgb]{ .886,  .937,  .855}79.22 & \cellcolor[rgb]{ .776,  .878,  .706}79.22 & \cellcolor[rgb]{ .776,  .878,  .706}81.12 & \cellcolor[rgb]{ .776,  .878,  .706}80.93 \\
    \midrule
    \textbf{Bushy\_Eyebrows} & \cellcolor[rgb]{ .988,  .894,  .839}53.58 & \cellcolor[rgb]{ 1,  .949,  .8}82 & \cellcolor[rgb]{ 1,  .949,  .8}84.97 & \cellcolor[rgb]{ 1,  .949,  .8}80 & \cellcolor[rgb]{ .886,  .937,  .855}80.73 & \cellcolor[rgb]{ .886,  .937,  .855}82.41 & \cellcolor[rgb]{ .776,  .878,  .706}82.41 & \cellcolor[rgb]{ .776,  .878,  .706}84.42 & \cellcolor[rgb]{ .776,  .878,  .706}84.26 \\
    \midrule
    \textbf{Chubby} & \cellcolor[rgb]{ .988,  .894,  .839}64.31 & \cellcolor[rgb]{ 1,  .949,  .8}73 & \cellcolor[rgb]{ 1,  .949,  .8}76.86 & \cellcolor[rgb]{ 1,  .949,  .8}75 & \cellcolor[rgb]{ .886,  .937,  .855}74.13 & \cellcolor[rgb]{ .886,  .937,  .855}75.19 & \cellcolor[rgb]{ .776,  .878,  .706}75.19 & \cellcolor[rgb]{ .776,  .878,  .706}76.13 & \cellcolor[rgb]{ .776,  .878,  .706}76.06 \\
    \midrule
    \textbf{Double\_Chin} & \cellcolor[rgb]{ .988,  .894,  .839}65.58 & \cellcolor[rgb]{ 1,  .949,  .8}78 & \cellcolor[rgb]{ 1,  .949,  .8}81.52 & \cellcolor[rgb]{ 1,  .949,  .8}78 & \cellcolor[rgb]{ .886,  .937,  .855}77.82 & \cellcolor[rgb]{ .886,  .937,  .855}79.19 & \cellcolor[rgb]{ .776,  .878,  .706}79.19 & \cellcolor[rgb]{ .776,  .878,  .706}80.76 & \cellcolor[rgb]{ .776,  .878,  .706}80.49 \\
    \midrule
    \textbf{Eyeglasses} & \cellcolor[rgb]{ .988,  .894,  .839}80.23 & \cellcolor[rgb]{ 1,  .949,  .8}95 & \cellcolor[rgb]{ 1,  .949,  .8}91.30 & \cellcolor[rgb]{ 1,  .949,  .8}92 & \cellcolor[rgb]{ .886,  .937,  .855}89.69 & \cellcolor[rgb]{ .886,  .937,  .855}90.76 & \cellcolor[rgb]{ .776,  .878,  .706}90.76 & \cellcolor[rgb]{ .776,  .878,  .706}91.72 & \cellcolor[rgb]{ .776,  .878,  .706}91.50 \\
    \midrule
    \textbf{Goatee} & \cellcolor[rgb]{ .988,  .894,  .839}77.41 & \cellcolor[rgb]{ 1,  .949,  .8}78 & \cellcolor[rgb]{ 1,  .949,  .8}82.97 & \cellcolor[rgb]{ 1,  .949,  .8}86 & \cellcolor[rgb]{ .886,  .937,  .855}81.72 & \cellcolor[rgb]{ .886,  .937,  .855}81.72 & \cellcolor[rgb]{ .776,  .878,  .706}81.72 & \cellcolor[rgb]{ .776,  .878,  .706}83.30 & \cellcolor[rgb]{ .776,  .878,  .706}83.01 \\
    \midrule
    \textbf{Gray\_Hair} & \cellcolor[rgb]{ .988,  .894,  .839}83.94 & \cellcolor[rgb]{ 1,  .949,  .8}84 & \cellcolor[rgb]{ 1,  .949,  .8}88.93 & \cellcolor[rgb]{ 1,  .949,  .8}88 & \cellcolor[rgb]{ .886,  .937,  .855}87.94 & \cellcolor[rgb]{ .886,  .937,  .855}87.94 & \cellcolor[rgb]{ .776,  .878,  .706}87.94 & \cellcolor[rgb]{ .776,  .878,  .706}88.37 & \cellcolor[rgb]{ .776,  .878,  .706}88.46 \\
    \midrule
    \textbf{Heavy\_Makeup} & \cellcolor[rgb]{ .988,  .894,  .839}87.21 & \cellcolor[rgb]{ 1,  .949,  .8}95 & \cellcolor[rgb]{ 1,  .949,  .8}95.85 & \cellcolor[rgb]{ 1,  .949,  .8}95 & \cellcolor[rgb]{ .886,  .937,  .855}94.80 & \cellcolor[rgb]{ .886,  .937,  .855}94.80 & \cellcolor[rgb]{ .776,  .878,  .706}94.80 & \cellcolor[rgb]{ .776,  .878,  .706}95.38 & \cellcolor[rgb]{ .776,  .878,  .706}95.39 \\
    \midrule
    \textbf{High\_Cheekbones} & \cellcolor[rgb]{ .988,  .894,  .839}63.34 & \cellcolor[rgb]{ 1,  .949,  .8}88 & \cellcolor[rgb]{ 1,  .949,  .8}88.38 & \cellcolor[rgb]{ 1,  .949,  .8}89 & \cellcolor[rgb]{ .886,  .937,  .855}86.53 & \cellcolor[rgb]{ .886,  .937,  .855}86.53 & \cellcolor[rgb]{ .776,  .878,  .706}86.53 & \cellcolor[rgb]{ .776,  .878,  .706}88.34 & \cellcolor[rgb]{ .776,  .878,  .706}88.34 \\
    \midrule
    \textbf{Male} & \cellcolor[rgb]{ .988,  .894,  .839}76.02 & \cellcolor[rgb]{ 1,  .949,  .8}94 & \cellcolor[rgb]{ 1,  .949,  .8}94.02 & \cellcolor[rgb]{ 1,  .949,  .8}93 & \cellcolor[rgb]{ .886,  .937,  .855}92.17 & \cellcolor[rgb]{ .886,  .937,  .855}92.17 & \cellcolor[rgb]{ .776,  .878,  .706}92.17 & \cellcolor[rgb]{ .776,  .878,  .706}92.81 & \cellcolor[rgb]{ .776,  .878,  .706}92.60 \\
    \midrule
    \textbf{Mouth\_Slightly\_Open} & \cellcolor[rgb]{ .988,  .894,  .839}57.02 & \cellcolor[rgb]{ 1,  .949,  .8}82 & \cellcolor[rgb]{ 1,  .949,  .8}83.51 & \cellcolor[rgb]{ 1,  .949,  .8}86 & \cellcolor[rgb]{ .886,  .937,  .855}79.03 & \cellcolor[rgb]{ .886,  .937,  .855}79.03 & \cellcolor[rgb]{ .776,  .878,  .706}79.03 & \cellcolor[rgb]{ .776,  .878,  .706}82.70 & \cellcolor[rgb]{ .776,  .878,  .706}82.50 \\
    \midrule
    \textbf{Mustache} & \cellcolor[rgb]{ .988,  .894,  .839}89.03 & \cellcolor[rgb]{ 1,  .949,  .8}92 & \cellcolor[rgb]{ 1,  .949,  .8}93.43 & \cellcolor[rgb]{ 1,  .949,  .8}95 & \cellcolor[rgb]{ .886,  .937,  .855}91.92 & \cellcolor[rgb]{ .886,  .937,  .855}91.92 & \cellcolor[rgb]{ .776,  .878,  .706}91.92 & \cellcolor[rgb]{ .776,  .878,  .706}93.27 & \cellcolor[rgb]{ .776,  .878,  .706}92.97 \\
    \midrule
    \textbf{Narrow\_Eyes} & \cellcolor[rgb]{ .988,  .894,  .839}63.45 & \cellcolor[rgb]{ 1,  .949,  .8}81 & \cellcolor[rgb]{ 1,  .949,  .8}82.86 & \cellcolor[rgb]{ 1,  .949,  .8}82 & \cellcolor[rgb]{ .886,  .937,  .855}78.94 & \cellcolor[rgb]{ .886,  .937,  .855}80.07 & \cellcolor[rgb]{ .776,  .878,  .706}80.07 & \cellcolor[rgb]{ .776,  .878,  .706}82.86 & \cellcolor[rgb]{ .776,  .878,  .706}82.75 \\
    \midrule
    \textbf{No\_Beard} & \cellcolor[rgb]{ .988,  .894,  .839}73.08 & \cellcolor[rgb]{ 1,  .949,  .8}79 & \cellcolor[rgb]{ 1,  .949,  .8}82.15 & \cellcolor[rgb]{ 1,  .949,  .8}81 & \cellcolor[rgb]{ .886,  .937,  .855}79.27 & \cellcolor[rgb]{ .886,  .937,  .855}79.27 & \cellcolor[rgb]{ .776,  .878,  .706}79.27 & \cellcolor[rgb]{ .776,  .878,  .706}80.65 & \cellcolor[rgb]{ .776,  .878,  .706}80.77 \\
    \midrule
    \textbf{Oval\_Face} & \cellcolor[rgb]{ .988,  .894,  .839}52.37 & \cellcolor[rgb]{ 1,  .949,  .8}74 & \cellcolor[rgb]{ 1,  .949,  .8}77.39 & \cellcolor[rgb]{ 1,  .949,  .8}75 & \cellcolor[rgb]{ .886,  .937,  .855}74.19 & \cellcolor[rgb]{ .886,  .937,  .855}74.19 & \cellcolor[rgb]{ .776,  .878,  .706}74.19 & \cellcolor[rgb]{ .776,  .878,  .706}76.51 & \cellcolor[rgb]{ .776,  .878,  .706}76.80 \\
    \midrule
    \textbf{Pale\_Skin} & \cellcolor[rgb]{ .988,  .894,  .839}50.82 & \cellcolor[rgb]{ 1,  .949,  .8}84 & \cellcolor[rgb]{ 1,  .949,  .8}93.32 & \cellcolor[rgb]{ 1,  .949,  .8}91 & \cellcolor[rgb]{ .886,  .937,  .855}88.36 & \cellcolor[rgb]{ .886,  .937,  .855}90.16 & \cellcolor[rgb]{ .776,  .878,  .706}90.16 & \cellcolor[rgb]{ .776,  .878,  .706}91.00 & \cellcolor[rgb]{ .776,  .878,  .706}90.97 \\
    \midrule
    \textbf{Pointy\_Nose} & \cellcolor[rgb]{ .988,  .894,  .839}68.4 & \cellcolor[rgb]{ 1,  .949,  .8}80 & \cellcolor[rgb]{ 1,  .949,  .8}84.14 & \cellcolor[rgb]{ 1,  .949,  .8}84 & \cellcolor[rgb]{ .886,  .937,  .855}81.50 & \cellcolor[rgb]{ .886,  .937,  .855}82.92 & \cellcolor[rgb]{ .776,  .878,  .706}82.92 & \cellcolor[rgb]{ .776,  .878,  .706}83.63 & \cellcolor[rgb]{ .776,  .878,  .706}84.20 \\
    \midrule
    \textbf{Receding\_Hairline} & \cellcolor[rgb]{ .988,  .894,  .839}56.36 & \cellcolor[rgb]{ 1,  .949,  .8}85 & \cellcolor[rgb]{ 1,  .949,  .8}86.25 & \cellcolor[rgb]{ 1,  .949,  .8}85 & \cellcolor[rgb]{ .886,  .937,  .855}83.91 & \cellcolor[rgb]{ .886,  .937,  .855}83.91 & \cellcolor[rgb]{ .776,  .878,  .706}83.91 & \cellcolor[rgb]{ .776,  .878,  .706}85.09 & \cellcolor[rgb]{ .776,  .878,  .706}84.90 \\
    \midrule
    \textbf{Rosy\_Cheeks} & \cellcolor[rgb]{ .988,  .894,  .839}81.46 & \cellcolor[rgb]{ 1,  .949,  .8}78 & \cellcolor[rgb]{ 1,  .949,  .8}87.92 & \cellcolor[rgb]{ 1,  .949,  .8}86 & \cellcolor[rgb]{ .886,  .937,  .855}85.55 & \cellcolor[rgb]{ .886,  .937,  .855}85.55 & \cellcolor[rgb]{ .776,  .878,  .706}85.55 & \cellcolor[rgb]{ .776,  .878,  .706}87.19 & \cellcolor[rgb]{ .776,  .878,  .706}87.08 \\
    \midrule
    \textbf{Sideburns} & \cellcolor[rgb]{ .988,  .894,  .839}69.38 & \cellcolor[rgb]{ 1,  .949,  .8}77 & \cellcolor[rgb]{ 1,  .949,  .8}83.13 & \cellcolor[rgb]{ 1,  .949,  .8}80 & \cellcolor[rgb]{ .886,  .937,  .855}79.42 & \cellcolor[rgb]{ .886,  .937,  .855}79.42 & \cellcolor[rgb]{ .776,  .878,  .706}79.42 & \cellcolor[rgb]{ .776,  .878,  .706}81.89 & \cellcolor[rgb]{ .776,  .878,  .706}81.76 \\
    \midrule
    \textbf{Smiling} & \cellcolor[rgb]{ .988,  .894,  .839}56.65 & \cellcolor[rgb]{ 1,  .949,  .8}91 & \cellcolor[rgb]{ 1,  .949,  .8}91.83 & \cellcolor[rgb]{ 1,  .949,  .8}92 & \cellcolor[rgb]{ .886,  .937,  .855}88.65 & \cellcolor[rgb]{ .886,  .937,  .855}88.65 & \cellcolor[rgb]{ .776,  .878,  .706}88.65 & \cellcolor[rgb]{ .776,  .878,  .706}90.77 & \cellcolor[rgb]{ .776,  .878,  .706}90.80 \\
    \midrule
    \textbf{Straight\_Hair} & \cellcolor[rgb]{ .988,  .894,  .839}60.1 & \cellcolor[rgb]{ 1,  .949,  .8}76 & \cellcolor[rgb]{ 1,  .949,  .8}78.53 & \cellcolor[rgb]{ 1,  .949,  .8}79 & \cellcolor[rgb]{ .886,  .937,  .855}77.09 & \cellcolor[rgb]{ .886,  .937,  .855}78.10 & \cellcolor[rgb]{ .776,  .878,  .706}78.10 & \cellcolor[rgb]{ .776,  .878,  .706}79.27 & \cellcolor[rgb]{ .776,  .878,  .706}78.91 \\
    \midrule
    \textbf{Wavy\_Hair} & \cellcolor[rgb]{ .988,  .894,  .839}57.94 & \cellcolor[rgb]{ 1,  .949,  .8}76 & \cellcolor[rgb]{ 1,  .949,  .8}81.61 & \cellcolor[rgb]{ 1,  .949,  .8}80 & \cellcolor[rgb]{ .886,  .937,  .855}77.02 & \cellcolor[rgb]{ .886,  .937,  .855}77.02 & \cellcolor[rgb]{ .776,  .878,  .706}77.02 & \cellcolor[rgb]{ .776,  .878,  .706}78.55 & \cellcolor[rgb]{ .776,  .878,  .706}78.28 \\
    \midrule
    \textbf{Wearing\_Earrings} & \cellcolor[rgb]{ .988,  .894,  .839}85.1 & \cellcolor[rgb]{ 1,  .949,  .8}94 & \cellcolor[rgb]{ 1,  .949,  .8}94.95 & \cellcolor[rgb]{ 1,  .949,  .8}94 & \cellcolor[rgb]{ .886,  .937,  .855}94.20 & \cellcolor[rgb]{ .886,  .937,  .855}94.20 & \cellcolor[rgb]{ .776,  .878,  .706}94.20 & \cellcolor[rgb]{ .776,  .878,  .706}94.59 & \cellcolor[rgb]{ .776,  .878,  .706}94.75 \\
    \midrule
    \textbf{Wearing\_Hat} & \cellcolor[rgb]{ .988,  .894,  .839}86.57 & \cellcolor[rgb]{ 1,  .949,  .8}88 & \cellcolor[rgb]{ 1,  .949,  .8}90.07 & \cellcolor[rgb]{ 1,  .949,  .8}92 & \cellcolor[rgb]{ .886,  .937,  .855}89.81 & \cellcolor[rgb]{ .886,  .937,  .855}90.23 & \cellcolor[rgb]{ .776,  .878,  .706}90.23 & \cellcolor[rgb]{ .776,  .878,  .706}90.25 & \cellcolor[rgb]{ .776,  .878,  .706}90.23 \\
    \midrule
    \textbf{Wearing\_Lipstick} & \cellcolor[rgb]{ .988,  .894,  .839}83.22 & \cellcolor[rgb]{ 1,  .949,  .8}95 & \cellcolor[rgb]{ 1,  .949,  .8}95.04 & \cellcolor[rgb]{ 1,  .949,  .8}93 & \cellcolor[rgb]{ .886,  .937,  .855}93.71 & \cellcolor[rgb]{ .886,  .937,  .855}93.71 & \cellcolor[rgb]{ .776,  .878,  .706}93.71 & \cellcolor[rgb]{ .776,  .878,  .706}94.07 & \cellcolor[rgb]{ .776,  .878,  .706}94.07 \\
    \midrule
    \textbf{Wearing\_Necklace} & \cellcolor[rgb]{ .988,  .894,  .839}78.54 & \cellcolor[rgb]{ 1,  .949,  .8}88 & \cellcolor[rgb]{ 1,  .949,  .8}89.94 & \cellcolor[rgb]{ 1,  .949,  .8}91 & \cellcolor[rgb]{ .886,  .937,  .855}88.71 & \cellcolor[rgb]{ .886,  .937,  .855}88.71 & \cellcolor[rgb]{ .776,  .878,  .706}88.71 & \cellcolor[rgb]{ .776,  .878,  .706}89.45 & \cellcolor[rgb]{ .776,  .878,  .706}89.59 \\
    \midrule
    \textbf{Wearing\_Necktie} & \cellcolor[rgb]{ .988,  .894,  .839}63.13 & \cellcolor[rgb]{ 1,  .949,  .8}79 & \cellcolor[rgb]{ 1,  .949,  .8}80.66 & \cellcolor[rgb]{ 1,  .949,  .8}81 & \cellcolor[rgb]{ .886,  .937,  .855}79.55 & \cellcolor[rgb]{ .886,  .937,  .855}79.55 & \cellcolor[rgb]{ .776,  .878,  .706}79.55 & \cellcolor[rgb]{ .776,  .878,  .706}81.70 & \cellcolor[rgb]{ .776,  .878,  .706}81.40 \\
    \midrule
    \textbf{Young} & \cellcolor[rgb]{ .988,  .894,  .839}78.59 & \cellcolor[rgb]{ 1,  .949,  .8}86 & \cellcolor[rgb]{ 1,  .949,  .8}85.84 & \cellcolor[rgb]{ 1,  .949,  .8}87 & \cellcolor[rgb]{ .886,  .937,  .855}83.90 & \cellcolor[rgb]{ .886,  .937,  .855}83.90 & \cellcolor[rgb]{ .776,  .878,  .706}83.90 & \cellcolor[rgb]{ .776,  .878,  .706}85.55 & \cellcolor[rgb]{ .776,  .878,  .706}85.68 \\
    \midrule
    \textbf{Mean Accuracy} & \cellcolor[rgb]{ .988,  .894,  .839}\textbf{71.27} & \cellcolor[rgb]{ 1,  .949,  .8}\textbf{83.85} & \cellcolor[rgb]{ 1,  .949,  .8}\textbf{86.31} & \cellcolor[rgb]{ 1,  .949,  .8}\textbf{86.15} & \cellcolor[rgb]{ .886,  .937,  .855}\textbf{84.02} & \cellcolor[rgb]{ .886,  .937,  .855}\textbf{84.36} & \cellcolor[rgb]{ .776,  .878,  .706}\textbf{84.36} & \cellcolor[rgb]{ .776,  .878,  .706}\textbf{85.85} & \cellcolor[rgb]{ .776,  .878,  .706}\textbf{85.82} \\
    \end{tabular}%
  \label{tab:CompareResultsLFWA}%
  }
\end{table*}%

%
%






\subsection{Cross-Dataset Testing Accuracies}
In table \ref{tab:CrossDataSetResults}, we presented the cross-dataset testing performances of AFFACT, DMTL and SPLITFACE (NSA product rule). For AFFACT and the proposed method, we presented two accuracies separated by $/$, the first one is for using the optimal threshold obtained from the validation set to find detection results (for AFFACT) or to normalize scores before applying product rule (for proposed). And, the second accuracy is obtained by using the mid value of the score range ($0$ for AFFACT which gives scores between $-1$ and $+1$, and $0.5$ for proposed method for which score is between $0$ and $1$) as the threshold. Higher accuracies are obtained by using optimal thresholds for the proposed method, while for AFFACT the accuracies dropped slightly. It can be seen from this table that, for cross-dataset testing (trained on CelebA and tested on LFWA or vice versa), the proposed method outperforms both AFFACT and DMTL with a relatively large margin. This again proves the generalization capability of SPLITFACE, which is achieved by the combination of its unique architecture with the committee machine. 

\begin{table}
  \centering
  \caption{Cross dataset results. The three numbers for each Train-Test pair are for AFFACT \cite{AFFECT_AlignmentFreeAttrib_Boult}, DMTL \cite{HuHan_DeepMultiTaskHeteroAttrib} and the Proposed method, respectively, from left to right.}
   \scalebox{0.85}{
    \begin{tabular}{c|ccc||ccc}
    \toprule
    \textbf{Train/Test} & \multicolumn{3}{c}{\textbf{CelebA}} & \multicolumn{3}{c}{\textbf{LFWA}} \\
    \midrule
    \textbf{CelebA} & 89.07/90.32 & 92.6  & 90.39/87.14 & 79.5/73.84 & 73    & 79.32/74.56 \\
    \midrule
    \midrule
    \textbf{LFWA} & -/- & 70.2  & 78.15/77.88 & -/- & 86    & 85.99/85.28 \\
    \bottomrule
    \end{tabular}%
    }
  \label{tab:CrossDataSetResults}%
\end{table}%

Next, we evaluated the performances SPLITFACE on the modified CelebA and LFWA partial face datasets. The results for evaluation on same and cross-dataset are presented in tables \ref{tab:CrossTableCelebAtoAll} and \ref{tab:CrossTableLFWAToAll}. In table \ref{tab:CrossTableCelebAtoAll}, results are presented form AFFACT and SPLITFACE network, both trained on the original CelebA training set and tested on the original and modified CelebA and LFWA datasets. The results for both using and not using optimal threshold (using $0$ for AFFACT and $0.5$ for SPLITFACE as threshold instead) are shown in the table. It can be seen that SPLITFACE, especially NSA with product rule and optimal threshold outperforms AFFACT in terms of accuracy for full face dataset, and both cross-domain and partial datasets. The differences are more prominent when using the optimal thresholds, which show that threshold normalization step with a piece-wise linear function can boost the overall performance. Similar scenario is found in Table \ref{tab:CrossTableLFWAToAll}, where SPLITFACE is trained on the original LFWA training set and tested on both original and partial CelebA and LFWA datasets. Since the no pre-trained version of AFFACT on LFWA is publicly available, the results for AFFACT could not be provided in this table. Note that in both tables \ref{tab:CrossTableCelebAtoAll} and \ref{tab:CrossTableLFWAToAll}, the committee machine approaches improves over the full face branches, especially for partial face datasets. This improvement can be attributed to the unique architecture of SPLITFACE that harnesses local information from unoccluded facial segments and to the ensemble aggregation approach by using committee machine.

\begin{table*}
  \centering
  \caption{Networks trained on CelebA and tested on both full and partial CelebA and LFWA datasets.}
    \scalebox{0.85}{
    \begin{tabular}{rcccccccc||ccccccc}
    \toprule
    \multicolumn{2}{c}{\textbf{Method}} & \textbf{CelebA} & \textbf{ C-U12} & \textbf{C-U34} & \textbf{C-L12} & \textbf{C-L34} & \textbf{C-R12} & \textbf{C-R34} & \textbf{LFWA} & \textbf{L-U12} & \textbf{L-U34} & \textbf{L-L12} & \textbf{L-L34} & \textbf{L-R12} & \textbf{L-R34} \\
    \midrule
          & \textbf{AFFACT} & \textbf{90.32} & 77.98 & 81.86 & 80.56 & 84.93 & 80.18 & 85.07 & 73.84 & \textbf{68.83} & 71.12 & 69.01 & 73    & 69.21 & \textbf{73.44} \\
\multicolumn{1}{c}{\textbf{Witout}} & \textbf{Full} & 86.76 & 80.99 & 84.34 & 83.86 & 85.39 & 83.45 & 85.71 & 73.52 & 67.28 & 70.25 & 69.94 & 72.69 & 70.01 & 72.67 \\
\multicolumn{1}{c}{\textbf{Optiamal }} & \textbf{HRP} & 86.93 & 81.46 & 84.51 & 84.23 & 85.97 & 84.3  & 86.29 & 73.54 & 67.51 & 70.51 & 70.08 & 72.13 & 70.38 & 72.64 \\
\multicolumn{1}{c}{\textbf{Threshold}} & \textbf{NSA Prod rule} & 87.14 & \textbf{83.24} & \textbf{85.53} & \textbf{84.6} & \textbf{86.79} & \textbf{84.84} & \textbf{86.5} & \textbf{74.56} & 68.61 & 71.45 & 70.34 & 73.2  & \textbf{70.35} & 73.32 \\
& \textbf{NSA Med rule} & 87.07 & 83.22 & 85.47 & 84.51 & 86.75 & 84.76 & 86.44 & 74.3  & 68.49 & \textbf{71.59} & \textbf{70.36} & \textbf{73.24} & 70.27 & 73.1 \\
\midrule
\midrule
    \multicolumn{1}{c}{} & \textbf{AFFACT} &  89.07     &      83.05 &  85.60    &    84.98  &     87.47  &   85.33    &  87.69     & 79.5  & 74.77 & 77.55 & 74.97 & 78.34 & 74.85 & 78.19 \\
    \multicolumn{1}{c}{\textbf{With}} & \textbf{Full} & 90.72 & 83.99 & 87.14 & 87.19 & 89.33 & 87.46 & 89.63 & 72.32 & 66.96 & 69.37 & 68.22 & 71.1  & 68.72 & 71.25 \\
    \multicolumn{1}{c}{\textbf{ Optimal }} & \textbf{HRP} & \textbf{90.8} & 84.27 & 87.59 & 87.11 & 89.31 & 87.82 & 89.71 & 72.67 & 67.28 & 69.94 & 67.94 & 70.91 & 68.42 & 71.35 \\
    \multicolumn{1}{c}{\textbf{Threshold}} & \textbf{NSA Prod rule} & 90.39 & 85.3  & \textbf{88.08} & \textbf{88.12} & \textbf{89.87} & \textbf{88.42} & \textbf{90.02} & 79.32 & \textbf{75.76} & 77.81 & \textbf{76.7} & \textbf{78.77} & \textbf{76.51} & \textbf{78.57} \\
          & \textbf{NSA Med rule} & 90.61 & \textbf{85.47} & 88.16 & 87.59 & 89.76 & 88.1  & 90.01 & \textbf{79.86} & 75.28 & \textbf{78.34} & 76.1  & 78.73 & 75.74 & 78.49 \\
    \bottomrule
    \end{tabular}%

  \label{tab:CrossTableCelebAtoAll}%
  }
\end{table*}%

\begin{table*}
  \centering
  \caption{Networks trained on LFWA and tested on both full and partial CelebA and LFWA datasets.}
    \scalebox{0.85}{
    \begin{tabular}{rcccccccc||ccccccc}
    \toprule
    \multicolumn{2}{c}{\textbf{Method}} & \textbf{CelebA} & \textbf{ C-U12} & \textbf{C-U34} & \textbf{C-L12} & \textbf{C-L34} & \textbf{C-R12} & \textbf{C-R34} & \textbf{LFWA} & \textbf{L-U12} & \textbf{L-U34} & \textbf{L-L12} & \textbf{L-L34} & \textbf{L-R12} & \textbf{L-R34} \\
    \midrule
    \multicolumn{1}{c}{\textbf{Without}} & \textbf{Full} & 75.87 & 64.84 & 71.67 & 70.93 & 73.67 & 70.77 & 74    & 83.52 & 67.11 & 74.76 & 75.08 & 79.74 & 75.8  & 80.24 \\
    \multicolumn{1}{c}{\textbf{Optimal }} & \textbf{HRP} & 75.94 & 65.03 & 71.94 & 70.78 & 73.64 & 70.58 & 74.02 & 83.84 & 67.59 & 75.36 & 74.93 & 79.81 & 75.7  & 80.45 \\
    \multicolumn{1}{c}{\textbf{Threshold}} & \textbf{NSA Prod rule} & 77.88 & \textbf{67.85} & \textbf{75.63} & \textbf{71.39} & \textbf{75.95} & \textbf{71.23} & \textbf{75.82} & \textbf{85.28} & \textbf{70.47} & \textbf{79.78} & \textbf{75.74} & \textbf{81.62} & \textbf{76.35} & \textbf{82.52} \\
          & \textbf{NSA Med rule} & \textbf{78.22} & 67.72 & 75.55 & 71.3  & 75.84 & 71.16 & 75.72 & 85.18 & 70.3  & 79.64 & 75.62 & 81.51 & 76.25 & 82.43 \\
\midrule
\midrule
    \multicolumn{1}{c}{\textbf{With}} & \textbf{Full} & 76.37 & 65.46 & 71.74 & 70.97 & 74    & 70.83 & 74.35 & 84.02 & 69.38 & 76.05 & 76.09 & 80.51 & 76.74 & 80.86 \\
    \multicolumn{1}{c}{\textbf{ Optimal }} & \textbf{HRP} & 76.58 & 66    & 72.18 & 71.22 & 74.21 & 71.16 & 74.61 & 84.36 & 70.13 & 76.66 & 76.42 & 80.85 & 77.06 & 81.28 \\
    \multicolumn{1}{c}{\textbf{Threshold}} & \textbf{NSA Prod rule} & \textbf{78.15} & \textbf{69.54} & \textbf{76.14} & \textbf{72.3} & \textbf{76.68} & \textbf{72.22} & \textbf{76.63} & \textbf{85.99} & \textbf{73.26} & \textbf{81.4} & \textbf{77.4} & \textbf{82.84} & \textbf{78.32} & \textbf{83.46} \\
          & \textbf{NSA Med rule} & 78.13 & 68.13 & 75.61 & 71.75 & 75.92 & 71.79 & 76.03 & 85.82 & 72.12 & 80.79 & 76.75 & 82.22 & 77.41 & 83.06 \\
    \bottomrule
    \end{tabular}%
  \label{tab:CrossTableLFWAToAll}%
  }
\end{table*}%

\subsection{Analysis of Performance Degradation with Occlusion}
\begin{figure*}
\centering
\includegraphics[width = 0.9\textwidth]{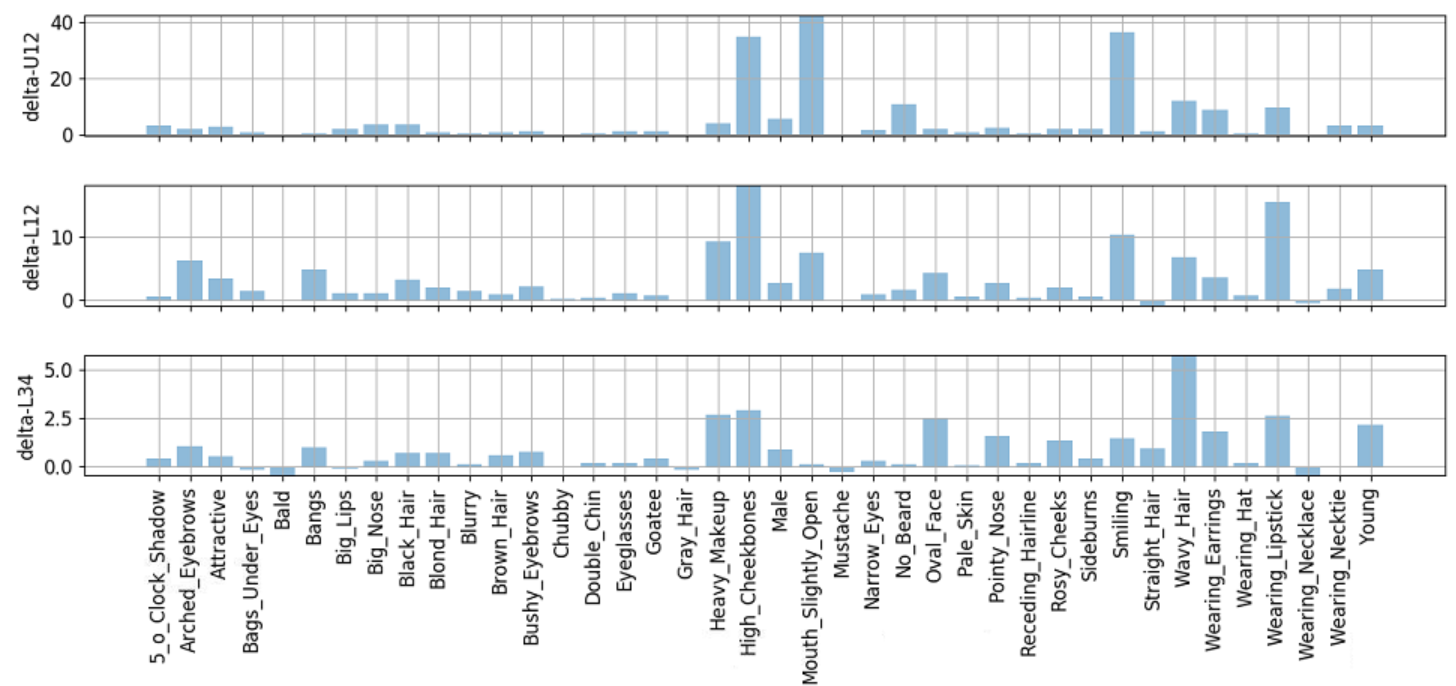}
\caption{Attribute-wise comparison of performance changes (w.r.to the performance on the unoccluded faces in CelebA) on the C-U12, C-L12 and C-L34 modified datasets. The vector of differences are denoted as delta-U12, delta-L12 and delta-L34 respectively.}
\label{PartialCompare}
\end{figure*}

One obvious observation from tables \ref{tab:CrossTableCelebAtoAll} and \ref{tab:CrossTableLFWAToAll} is that the attribute detection accuracies decrease with increasing occlusion. For example, all the methods achieve higher accuracies for upper three-fourth faces present in C-U34 and L-U34 in comparison to C-U12 and L-U12, respectively. In this section, we explore the effect of occlusion on the accuracy of each attribute using Fig \ref{PartialCompare} which plots the decrease in accuracy of SPLITFACE (after stage 1 before output pruning) for the partial CelebA datasets, C$-P, P \in \{\mbox{U12, L12, L34}\}$ described in section \ref{datasets} with respect to full face accuracy. The differences are denoted as delta-U12, delta-L12 and delta-L34, respectively. We observe that SPLITFACE fails in C-L12 and C-L34 for the same attributes such as wavy hair, high cheekbones and wearing lipstick, since part of the right side of the faces containing vital information in these regard are occluded in both cases. On the other hand, C-U12 has reduced performance for attributes like `mouth slightly open', `no beard' and `smiling', which are attributes localized in the lower part of the face, which is not visible in C-U12. So, SPLITFACE avoids catastrophic failures during occlusion, since the prediction accuracy of other attributes remain near constant and only invisible localized attributes' performance degrade. The output pruning step of SPLITFACE removes the attributes for which a segment performs badly in the first stage. When trained, SPLITFACE utilizes information from different segments to bolster its decision about an attribute as well as fill up the gaps in attributes in one segment by using information from other segments which predict those missing attributes.










\subsection{Performance for Partial Face Augmentation}
We also trained SPLITFACE with training samples from the modified partial face datasets. When training, samples from the modified datasets were picked with a $0.3$ probability in each batch while the rest of the samples came from the original datasets. The performances of the networks trained in this manner are presented in tables \ref{tab:CelebAWithFS} and \ref{tab:LFWAWithFS}. In comparison to tables \ref{tab:CrossTableCelebAtoAll} and \ref{tab:CrossTableLFWAToAll}, we can see that for both partially modified CelebA and LFWA datasets, the performances improved greatly when the partial faces are augment the training samples in addition to segment dropout.

\begin{table*}
  \centering
  \caption{Performance of SPLITFACE trained on original and modified CelebA (70-30 ratio).}
    \begin{tabular}{crcccccc}
    \toprule
    \textbf{Methods} & \textbf{CelebA} & \textbf{C-U12} & \textbf{C-U34} & \textbf{C-L12} & \textbf{C-L34} & \textbf{C-R12} & \textbf{C-R34} \\
    \midrule
    \textbf{Full } & \multicolumn{1}{c}{90.42} & 88.01 & 89.7  & 89.56 & 90.02 & 89.54 & 90.07 \\
    \textbf{HRP} & \multicolumn{1}{c}{90.18} & 87.94 & 89.55 & 89.23 & 89.77 & 89.21 & 89.71 \\
    \textbf{NSA Prod rule} & \multicolumn{1}{c}{90.39} & 88.43 & 89.77 & 89.76 & 90.02 & \textbf{89.86} & 90.11 \\
    \textbf{NSA Med rule} & \multicolumn{1}{c}{\textbf{90.52}} & \textbf{88.46} & \textbf{90.05} & \textbf{89.85} & \textbf{90.16} & 89.85 & \textbf{90.21} \\
    \bottomrule
    \end{tabular}%
  \label{tab:CelebAWithFS}%
\end{table*}%

\begin{table*}[htbp]
  \centering
  \caption{Performance of SPLITFACE trained on original and modified LFWA (70-30 ratio).}
    \begin{tabular}{cccccccc}
    \toprule
    \textbf{Methods} & \textbf{LFWA} & \textbf{L-U12} & \textbf{L-U34} & \textbf{L-L12} & \textbf{L-L34} & \textbf{L-R12} & \textbf{L-R34} \\
    \midrule
    Full  & 83.93 & 80.31 & 83.01 & 81.6  & 83    & 81.93 & 83.37 \\
    HRP   & 85.46 & \textbf{82.39} & 84.45 & \textbf{83.36} & 84.59 & \textbf{83.56} & 84.79 \\
    NSA Prod rule & \textbf{86.04} & 82.06 & 84.87 & 83.07 & \textbf{84.97} & 83.43 & 85.14 \\
    NSA Med rule & 85.97 & 82.17 & \textbf{84.88} & 83.06 & 84.87 & 83.4  & \textbf{85.23} \\
    \bottomrule
    \end{tabular}%
  \label{tab:LFWAWithFS}%
\end{table*}%

\section{Conclusion and Future Work}\label{Conclusion}
In this paper, we introduced SPLITFACE, an algorithm for facial attribute extraction utilizing multiple facial segments, a unique deep convolutional network, and a committee machine approach for ensemble aggregation. Through extensive experimentation, we have shown that the proposed method outperforms state-of-the-art facial attribute extraction methods when the faces are partially visible. Also, utilizing a committee machine approach, SPLITFACE achieved better generalization and therefore superior performance across domains. Moreover, when trained with both segment dropout and partial face data, the network achieved even higher attribute detection accuracy for partially visible faces. 
 The overall accuracies might be boosted by replacing the full face and segment branches with more advanced deep neural network architecture such as ResNet. On the other hand, since the segments are heavily overlapping and therefore can assist each other greatly, similar performance might be achievable with smaller input images. Finally, it would be interesting to see if a cross-stitch network \cite{crossstitchNet} can improve performance when connected to the segment network branches at certain intervals, by allowing the segment networks to share data. 

\ifCLASSOPTIONcompsoc
  \section*{Acknowledgments}
\else
  \section*{Acknowledgment}
\fi

This research is based upon work supported by the Office of  the  Director  of  National  Intelligence  (ODNI),  Intelligence  Advanced  Research  Projects  Activity  (IARPA),  via IARPA  R$\&$D  Contract  No.  2014-14071600012.  The  views and  conclusions  contained  herein  are  those  of  the  authors and should not be interpreted as necessarily representing the official policies or endorsements, either expressed or implied, of  the  ODNI,  IARPA,  or  the  U.S.  Government.  The  U.S.
Government is authorized to reproduce and distribute reprints for  Governmental  purposes  notwithstanding  any  copyright annotation thereon.

\bibliographystyle{IEEEtran}
\bibliography{FSAD}

\begin{thebibliography}{10}
\providecommand{\url}[1]{#1}
\csname url@samestyle\endcsname
\providecommand{\newblock}{\relax}
\providecommand{\bibinfo}[2]{#2}
\providecommand{\BIBentrySTDinterwordspacing}{\spaceskip=0pt\relax}
\providecommand{\BIBentryALTinterwordstretchfactor}{4}
\providecommand{\BIBentryALTinterwordspacing}{\spaceskip=\fontdimen2\font plus
\BIBentryALTinterwordstretchfactor\fontdimen3\font minus
  \fontdimen4\font\relax}
\providecommand{\BIBforeignlanguage}[2]{{%
\expandafter\ifx\csname l@#1\endcsname\relax
\typeout{** WARNING: IEEEtran.bst: No hyphenation pattern has been}%
\typeout{** loaded for the language `#1'. Using the pattern for}%
\typeout{** the default language instead.}%
\else
\language=\csname l@#1\endcsname
\fi
#2}}
\providecommand{\BIBdecl}{\relax}
\BIBdecl

\bibitem{JointFaceAndAttribDetect_He_2017}
\BIBentryALTinterwordspacing
K.~He, Y.~Fu, and X.~Xue, ``A jointly learned deep architecture for facial
  attribute analysis and face detection in the wild,'' \emph{CoRR}, vol.
  abs/1707.08705, 2017. [Online]. Available:
  \url{http://arxiv.org/abs/1707.08705}
\BIBentrySTDinterwordspacing

\bibitem{MOON_Rudd2016}
\BIBentryALTinterwordspacing
E.~M. Rudd, M.~G{\"u}nther, and T.~E. Boult, \emph{MOON: A Mixed Objective
  Optimization Network for the Recognition of Facial Attributes}.\hskip 1em
  plus 0.5em minus 0.4em\relax Cham: Springer International Publishing, 2016,
  pp. 19--35. [Online]. Available:
  \url{https://doi.org/10.1007/978-3-319-46454-1_2}
\BIBentrySTDinterwordspacing

\bibitem{EmilyNet_AAI_Multitask}
\BIBentryALTinterwordspacing
E.~Hand and R.~Chellappa, ``Attributes for improved attributes: A multi-task
  network utilizing implicit and explicit relationships for facial attribute
  classification,'' 2017. [Online]. Available:
  \url{https://aaai.org/ocs/index.php/AAAI/AAAI17/paper/view/14749}
\BIBentrySTDinterwordspacing

\bibitem{HuHan_DeepMultiTaskHeteroAttrib}
\BIBentryALTinterwordspacing
H.~Han, A.~K. Jain, S.~Shan, and X.~Chen, ``Heterogeneous face attribute
  estimation: {A} deep multi-task learning approach,'' \emph{CoRR}, vol.
  abs/1706.00906, 2017. [Online]. Available:
  \url{http://arxiv.org/abs/1706.00906}
\BIBentrySTDinterwordspacing

\bibitem{KumarBelhumer_FaceVerificationFromAttribute2009}
N.~Kumar, A.~Berg, P.~Belhumeur, and S.~Nayar, \emph{Attribute and simile
  classifiers for face verification}, 2009, pp. 365--372.

\bibitem{KumarBelhumeurImagesearch2011}
N.~Kumar, A.~Berg, P.~N. Belhumeur, and S.~Nayar, ``Describable visual
  attributes for face verification and image search,'' \emph{IEEE Transactions
  on Pattern Analysis and Machine Intelligence}, vol.~33, no.~10, pp.
  1962--1977, Oct 2011.

\bibitem{videoSurveillance2009}
D.~A. Vaquero, R.~S. Feris, D.~Tran, L.~Brown, A.~Hampapur, and M.~Turk,
  ``Attribute-based people search in surveillance environments,'' in \emph{2009
  Workshop on Applications of Computer Vision (WACV)}, Dec 2009, pp. 1--8.

\bibitem{VMP_SPM_AA_2016}
V.~M. Patel, R.~Chellappa, D.~Chandra, and B.~Barbello, ``Continuous user
  authentication on mobile devices: Recent progress and remaining challenges,''
  \emph{IEEE Signal Processing Magazine}, vol.~33, no.~4, pp. 49--61, July
  2016.

\bibitem{SegFaceDeepSegFace_FG2017}
U.~Mahbub, S.~Sarkar, and R.~Chellappa, ``Pooling facial segments to face: The
  shallow and deep ends,'' in \emph{12th IEEE International Conference on
  Automatic Face and Gesture Recognition (FG 2017)}, May 2017.

\bibitem{Huo2016DeepAD}
Z.-W. Huo, X.~Yang, C.~Xing, Y.~Zhou, P.~Hou, J.~Lv, and X.~Geng, ``Deep age
  distribution learning for apparent age estimation,'' \emph{2016 IEEE
  Conference on Computer Vision and Pattern Recognition Workshops (CVPRW)}, pp.
  722--729, 2016.

\bibitem{AgeAndGender_Eidinger}
E.~Eidinger, R.~Enbar, and T.~Hassner, ``Age and gender estimation of
  unfiltered faces,'' \emph{IEEE Transactions on Information Forensics and
  Security}, vol.~9, no.~12, pp. 2170--2179, Dec 2014.

\bibitem{JoingAgeGenderEthnicity_Guo_FGNonCNN}
G.~Guo and G.~Mu, ``Joint estimation of age, gender and ethnicity: Cca vs.
  pls,'' in \emph{2013 10th IEEE International Conference and Workshops on
  Automatic Face and Gesture Recognition (FG)}, April 2013, pp. 1--6.

\bibitem{CelebA_liu2015faceattributes}
Z.~Liu, P.~Luo, X.~Wang, and X.~Tang, ``Deep learning face attributes in the
  wild,'' in \emph{Proceedings of International Conference on Computer Vision
  (ICCV)}, 2015.

\bibitem{FasterRCNN_NIPS2015}
\BIBentryALTinterwordspacing
S.~Ren, K.~He, R.~Girshick, and J.~Sun, ``Faster r-cnn: Towards real-time
  object detection with region proposal networks,'' in \emph{Advances in Neural
  Information Processing Systems 28}, C.~Cortes, N.~D. Lawrence, D.~D. Lee,
  M.~Sugiyama, and R.~Garnett, Eds.\hskip 1em plus 0.5em minus 0.4em\relax
  Curran Associates, Inc., 2015, pp. 91--99. [Online]. Available:
  \url{\url{http://papers.nips.cc/paper/5638-faster-r-cnn-towards-real-time-object-detection-with-region-proposal-networks.pdf}}
\BIBentrySTDinterwordspacing

\bibitem{DRUID_Umahbub}
\BIBentryALTinterwordspacing
U.~Mahbub, S.~Sarkar, and R.~Chellappa, ``Partial face detection in the mobile
  domain,'' \emph{CoRR}, vol. abs/1704.02117, 2017. [Online]. Available:
  \url{http://arxiv.org/abs/1704.02117}
\BIBentrySTDinterwordspacing

\bibitem{UnalignedFA_HuiDing}
\BIBentryALTinterwordspacing
H.~Ding, H.~Zhou, S.~K. Zhou, and R.~Chellappa, ``A deep cascade network for
  unaligned face attribute classification,'' \emph{CoRR}, vol. abs/1709.03851,
  2017. [Online]. Available: \url{http://arxiv.org/abs/1709.03851}
\BIBentrySTDinterwordspacing

\bibitem{AFFECT_AlignmentFreeAttrib_Boult}
\BIBentryALTinterwordspacing
M.~G{\"{u}}nther, A.~Rozsa, and T.~E. Boult, ``{AFFACT} - alignment free facial
  attribute classification technique,'' \emph{CoRR}, vol. abs/1611.06158, 2016.
  [Online]. Available: \url{http://arxiv.org/abs/1611.06158}
\BIBentrySTDinterwordspacing

\bibitem{FSFD_Mahbub}
U.~Mahbub, V.~M. Patel, D.~Chandra, B.~Barbello, and R.~Chellappa, ``Partial
  face detection for continuous authentication,'' in \emph{2016 IEEE
  International Conference on Image Processing (ICIP)}, Sept 2016, pp.
  2991--2995.

\bibitem{UltraFace_RRanjan}
R.~Ranjan, S.~Sankaranarayanan, C.~D. Castillo, and R.~Chellappa, ``An
  all-in-one convolutional neural network for face analysis,'' in \emph{2017
  12th IEEE International Conference on Automatic Face Gesture Recognition (FG
  2017)}, May 2017, pp. 17--24.

\bibitem{Zhou_2016_CVPR}
B.~Zhou, A.~Khosla, A.~Lapedriza, A.~Oliva, and A.~Torralba, ``Learning deep
  features for discriminative localization,'' in \emph{The IEEE Conference on
  Computer Vision and Pattern Recognition (CVPR)}, June 2016.

\bibitem{7961801}
U.~Mahbub, S.~Sarkar, and R.~Chellappa, ``Pooling facial segments to face: The
  shallow and deep ends,'' in \emph{2017 12th IEEE International Conference on
  Automatic Face Gesture Recognition (FG 2017)}, May 2017, pp. 634--641.

\bibitem{NIN}
\BIBentryALTinterwordspacing
M.~Lin, Q.~Chen, and S.~Yan, ``Network in network,'' \emph{CoRR}, vol.
  abs/1312.4400, 2013. [Online]. Available:
  \url{http://arxiv.org/abs/1312.4400}
\BIBentrySTDinterwordspacing

\bibitem{Cappelli_ThresholdNorm}
\BIBentryALTinterwordspacing
R.~Cappelli, D.~Maio, and D.~Maltoni, ``Combining fingerprint classifiers,'' in
  \emph{Proceedings of the First International Workshop on Multiple Classifier
  Systems}, ser. MCS '00.\hskip 1em plus 0.5em minus 0.4em\relax London, UK,
  UK: Springer-Verlag, 2000, pp. 351--361. [Online]. Available:
  \url{http://dl.acm.org/citation.cfm?id=648054.746359}
\BIBentrySTDinterwordspacing

\bibitem{Jain_CombinationApproach}
\BIBentryALTinterwordspacing
A.~Jain, K.~Nandakumar, and A.~Ross, ``Score normalization in multimodal
  biometric systems,'' \emph{Pattern Recogn.}, vol.~38, no.~12, pp. 2270--2285,
  Dec. 2005. [Online]. Available:
  \url{http://dx.doi.org/10.1016/j.patcog.2005.01.012}
\BIBentrySTDinterwordspacing

\bibitem{DBLP:journals/corr/cs-NE-9905012}
\BIBentryALTinterwordspacing
K.~Tumer and J.~Ghosh, ``Linear and order statistics combiners for pattern
  classification,'' \emph{CoRR}, vol. cs.NE/9905012, 1999. [Online]. Available:
  \url{http://arxiv.org/abs/cs.NE/9905012}
\BIBentrySTDinterwordspacing

\bibitem{ADAMOptimizer}
\BIBentryALTinterwordspacing
D.~P. Kingma and J.~Ba, ``Adam: {A} method for stochastic optimization,''
  \emph{CoRR}, vol. abs/1412.6980, 2014. [Online]. Available:
  \url{http://arxiv.org/abs/1412.6980}
\BIBentrySTDinterwordspacing

\bibitem{KERAS_chollet2015keras}
F.~Chollet \emph{et~al.}, ``Keras,'' \url{https://github.com/keras-team/keras},
  2015.

\bibitem{Tensorflow}
\BIBentryALTinterwordspacing
M.~Abadi, A.~Agarwal, P.~Barham, E.~Brevdo, Z.~Chen, C.~Citro, G.~S. Corrado,
  A.~Davis, J.~Dean, M.~Devin, S.~Ghemawat, I.~J. Goodfellow, A.~Harp,
  G.~Irving, M.~Isard, Y.~Jia, R.~J{\'{o}}zefowicz, L.~Kaiser, M.~Kudlur,
  J.~Levenberg, D.~Man{\'{e}}, R.~Monga, S.~Moore, D.~G. Murray, C.~Olah,
  M.~Schuster, J.~Shlens, B.~Steiner, I.~Sutskever, K.~Talwar, P.~A. Tucker,
  V.~Vanhoucke, V.~Vasudevan, F.~B. Vi{\'{e}}gas, O.~Vinyals, P.~Warden,
  M.~Wattenberg, M.~Wicke, Y.~Yu, and X.~Zheng, ``Tensorflow: Large-scale
  machine learning on heterogeneous distributed systems,'' \emph{CoRR}, vol.
  abs/1603.04467, 2016. [Online]. Available:
  \url{http://arxiv.org/abs/1603.04467}
\BIBentrySTDinterwordspacing

\bibitem{crossstitchNet}
\BIBentryALTinterwordspacing
I.~Misra, A.~Shrivastava, A.~Gupta, and M.~Hebert, ``Cross-stitch networks for
  multi-task learning,'' \emph{CoRR}, vol. abs/1604.03539, 2016. [Online].
  Available: \url{http://arxiv.org/abs/1604.03539}
\BIBentrySTDinterwordspacing

\end{thebibliography}

\begin{IEEEbiography}
[{\includegraphics[width=1in,height=1.25in,clip,keepaspectratio]{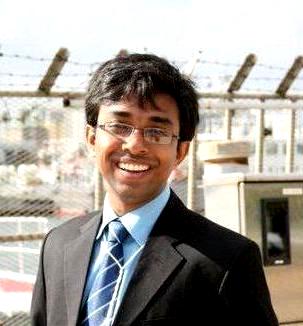}}]{Upal Mahbub} received the M.Sc. degree in Electrical and Computer Engineering from the University of Maryland College Park in 2017 and the B.Sc. and M.Sc. degree in Electrical and Electronic Engineering from the Department of EEE, BUET, Dhaka, Bangladesh, in 2009 and 2011, respectively. He currently a doctoral candidate in the department of ECE at the University of Maryland, College Park. His doctoral research is primarily focused on multi-modal active user authentication on smartphones using computer vision and machine learning techniques. Upal is the recipient of the best paper award at IEEE UEMCON 2016, best poster award at BTAS 2016, best paper award at ICCIT 2011 and distinguished graduate fellowship from the A. James Clark School of Engineering at the University of Maryland.
\end{IEEEbiography}

\begin{IEEEbiography}[{\includegraphics[width=1in,height=1.25in,clip,keepaspectratio]{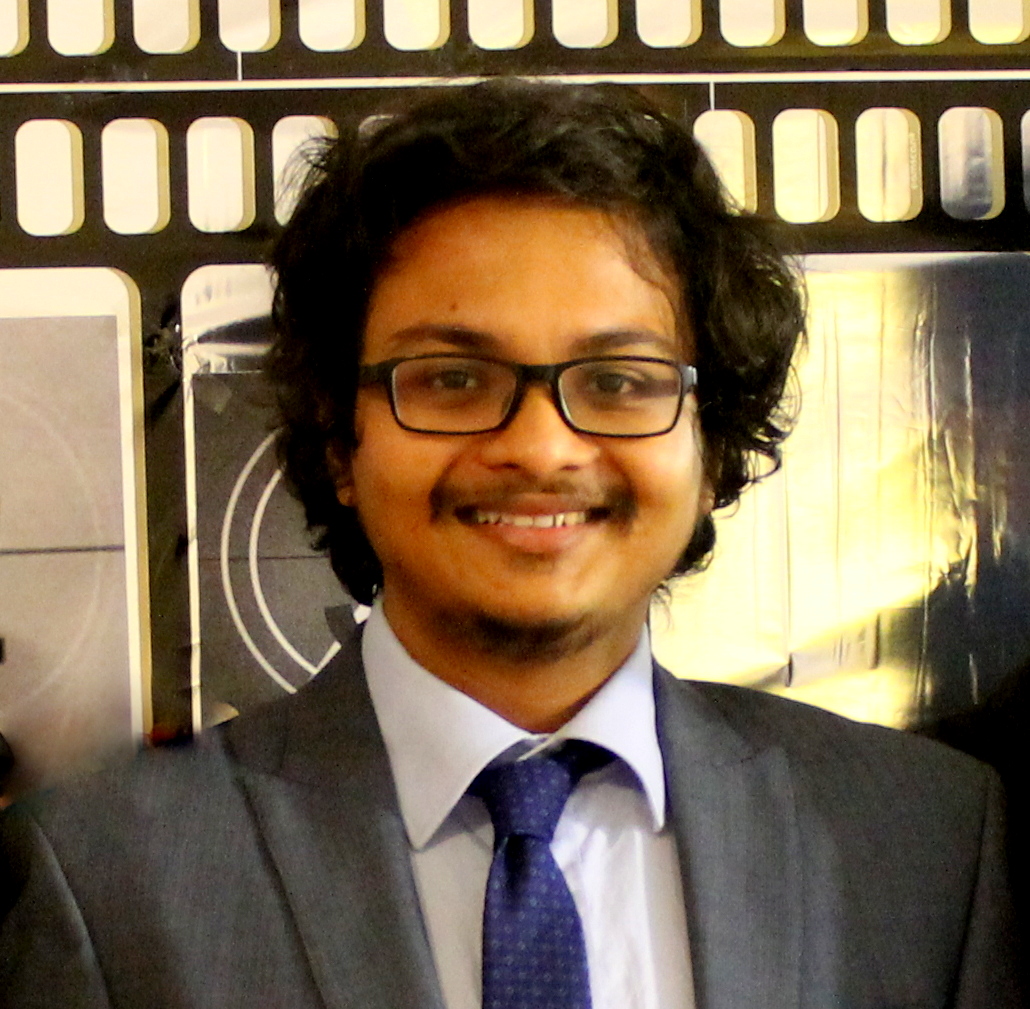}}]{Sayantan Sarkar} received his B.Tech. degree in Instrumentation Engineering from Indian Institute of Technology Kharagpur, India, in 2012. After working in mobile computing at Nvidia for 2 years, he is currently a Research Assistant at University of Maryland College Park. His research interests include face analysis and active authentication using deep learning. He received Best Poster Award at IEEE BTAS 2016.
\end{IEEEbiography}


\begin{IEEEbiography}[{\includegraphics[width=1in,height=1.25in,clip,keepaspectratio]{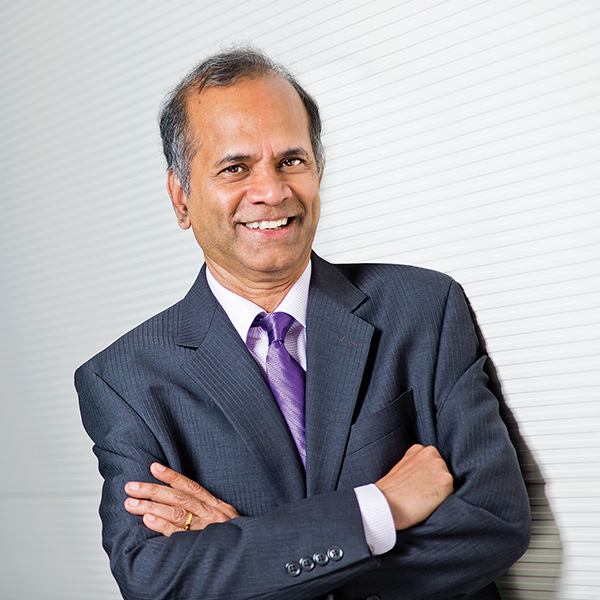}}]{Rama Chellappa} is a Distinguished University Professor, a Minta Martin Professor in Engineering and Chair of the Department of Electrical and Computer Engineering at the University of Maryland, College Park, MD. He received the B.E. (Hons.) degree in Electronics and Communication Engineering from the University of Madras, India and the M.E. (with Distinction) degree from the Indian Institute of Science, Bangalore, India. He received the M.S.E.E. and Ph.D. Degrees in Electrical Engineering from Purdue University, West Lafayette, IN. At UMD, he is an affiliate Professor of Computer Science Department, Applied Mathematics and Scientific Computing Program, member of the Center for Automation Research and a Permanent Member of the Institute for Advanced Computer Studies.  His current research interests span many areas in image processing, computer vision and machine learning. Prof. Chellappa is a recipient of an NSF Presidential Young Investigator Award and four IBM Faculty Development Awards. He received two paper awards and the K.S. Fu Prize from the International Association of Pattern Recognition (IAPR). He is a recipient of the Society, Technical Achievement and Meritorious Service Awards from the IEEE Signal Processing Society. He also received the Technical Achievement and Meritorious Service Awards from the IEEE Computer Society. Recently, he received the inaugural Leadership Award from the IEEE Biometrics Council. At UMD, he has received numerous college and university level recognitions for research, teaching, innovation and mentoring of undergraduate students. In 2010, he was recognized as an Outstanding ECE by Purdue University. He received the Distinguished Alumni Award from the Indian Institute of Science in 2016. Prof. Chellappa served as the EIC of IEEE Transactions on Pattern Analysis and Machine Intelligence, as the Co-EIC of Graphical Models and Image Processing, as the Associate Editor of four IEEE Transactions, as a Co-Guest Editor of many special issues, and is currently on the Editorial Board of SIAM Jl. of Imaging Science and Image and Vision Computing.  He has also served as the General and Technical Program Chair/Co-Chair for several IEEE international and national conferences and workshops. He is a Golden Core Member of the IEEE Computer Society, served as a Distinguished Lecturer of the IEEE Signal Processing Society and as the President of IEEE Biometrics Council. He is a Fellow of IEEE, IAPR, OSA, AAAS, ACM, and AAAI and holds six patents.
\end{IEEEbiography}

\end{document}